\documentclass{article} %
\usepackage{iclr2024_conference,times}

\usepackage{amsmath,amsfonts,bm}

\def\eqref#1{equation~\ref{#1}}

\def\1{\bm{1}}

\DeclareMathAlphabet{\mathsfit}{\encodingdefault}{\sfdefault}{m}{sl}
\SetMathAlphabet{\mathsfit}{bold}{\encodingdefault}{\sfdefault}{bx}{n}

\usepackage{url}
\usepackage{booktabs}       %
\usepackage{amsfonts}       %
\usepackage{nicefrac}       %
\usepackage{microtype}      %
\usepackage{amsmath,amsthm}
\usepackage{graphicx}
\usepackage{wrapfig}
\usepackage[font=small]{caption} 
\usepackage{listings}
\usepackage{algorithm}
\usepackage[noend]{algorithmic}
\usepackage{xspace} 
\usepackage{bbm}

\usepackage[toc,page,header]{appendix}
\usepackage{minitoc}
\usepackage{xpatch}

\usepackage{subcaption}
\usepackage{multirow, makecell}
\usepackage{multicol}
\usepackage{array}
\usepackage{amssymb,bm,amsthm}
\usepackage{enumitem}
\usepackage[misc,geometry]{ifsym}
\usepackage{adjustbox}
\usepackage{xfrac}
\usepackage{listings}
\usepackage{ltablex}  %
\usepackage{tabularx}
\definecolor{codegreen}{rgb}{0,0.6,0}
\definecolor{codegray}{rgb}{0.5,0.5,0.5}
\definecolor{codepink}{RGB}{252, 142, 172}
\definecolor{codepurple}{rgb}{0.58,0,0.82}
\definecolor{backcolour}{RGB}{245,245,245}
\lstdefinestyle{mystyle}{
    backgroundcolor=\color{backcolour},   
    commentstyle=\color{magenta},
    keywordstyle=\color{blue},
    numberstyle=\tiny\color{codegray},
    stringstyle=\color{codepurple},
    basicstyle=\fontfamily{\ttdefault}\footnotesize,
    breakatwhitespace=false,         
    breaklines=true,                 
    keepspaces=true,    
    frame=single,
    numbersep=5pt,                  
    showspaces=false,                
    showstringspaces=false,
    showtabs=false,                  
    tabsize=2,
    classoffset=1, %
    keywordstyle=\color{violet},
    classoffset=0,
}
\usepackage{color}
\usepackage{xcolor}
\definecolor{citecolor}{HTML}{3498DB}
\definecolor{linkcolor}{HTML}{E74C3C}
\usepackage[pagebackref=false,breaklinks=true,colorlinks,bookmarks=false,citecolor=citecolor,linkcolor=black]{hyperref}

\lstdefinelanguage{JavaScript}{
  keywords={typeof, new, true, false, catch, function, return, null, catch, switch, var, if, in, while, do, else, case, break},
  keywordstyle=\color{blue}\bfseries,
  ndkeywords={class, export, boolean, throw, implements, import, this},
  ndkeywordstyle=\color{darkgray}\bfseries,
  identifierstyle=\color{black},
  sensitive=false,
  comment=[l]{//},
  morecomment=[s]{/*}{*/},
  commentstyle=\color{purple}\ttfamily,
  stringstyle=\color{red}\ttfamily,
  morestring=[b]',
  morestring=[b]"
}

\usepackage{tabularx}
\newcolumntype{L}[1]{>{\raggedright\let\newline\\\arraybackslash\hspace{0pt}}m{#1}}
\newcolumntype{C}[1]{>{\centering\let\newline\\\arraybackslash\hspace{0pt}}m{#1}}
\newcolumntype{R}[1]{>{\raggedleft\let\newline\\\arraybackslash\hspace{0pt}}m{#1}}
\newcolumntype{Y}{>{\centering\arraybackslash}X}

\usepackage{soul}

\definecolor{snsgray}{RGB}{179, 179, 179}
\definecolor{snsorange}{RGB}{252, 141, 98}
\definecolor{snsblue}{RGB}{141, 160, 203}

\definecolor{coolgrey}{RGB}{157,157,157}
\definecolor{lightgrey}{RGB}{235,238,238}
\definecolor{lightteal}{RGB}{198,211,222}
\definecolor{cyan}{RGB}{136, 204, 238}
\definecolor{teal}{RGB}{68, 170, 153}
\definecolor{sand}{RGB}{221, 204, 119}
\definecolor{rose}{RGB}{204, 102, 119}
\definecolor{red}{RGB}{250, 94, 91}
\definecolor{orange}{RGB}{255, 200, 63}
\definecolor{yellow}{RGB}{254, 239, 109}

\definecolor{darkred}{RGB}{255, 184, 119}
\definecolor{darkgreen}{rgb}{0.09, 0.45, 0.27}

\newcommand{\arxiv}[1]{\textcolor{black}{#1}}

\newtheorem{theorem}{Theorem}[section]

\theoremstyle{definition}
\newtheorem{definition}[theorem]{Definition}

\usepackage[toc,page,header]{appendix}
\usepackage{minitoc}

\title{\ourmethod: Human-level Reward Design via \\ Coding Large Language Models}

\newcommand{\weburl}{\url{https://eureka-research.github.io}}

\author{%
    Yecheng Jason Ma$^{1\,2 \text{ \Letter}}$ , William Liang$^2$, Guanzhi Wang$^{1\,3}$, De-An Huang$^{1}$, Osbert Bastani$^{2}$,\\
    \textbf{Dinesh Jayaraman$^{2}$, Yuke Zhu$^{1\,4}$, Linxi ``Jim'' Fan$^{1\text{ \Letter}\,\dag}$, Anima Anandkumar$^{1\,3\,\dag }$}\\
    \hspace{2.3cm} $^1$NVIDIA, $^2$UPenn, $^3$Caltech, $^4$UT Austin; $^\dag$Equal advising\\
    \hspace{3cm} \weburl
}

\newcommand{\ourmethod}{\mbox{\textsc{Eureka}}\xspace}

\iclrfinalcopy %
\begin{document}
\doparttoc %
\faketableofcontents %

\maketitle

\begin{abstract}

Large Language Models (LLMs) have excelled as high-level semantic planners for sequential decision-making tasks. However, harnessing them to learn complex low-level manipulation tasks, such as dexterous pen spinning, remains an open problem. We bridge this fundamental gap and present \ourmethod, a human-level reward design algorithm powered by LLMs. \ourmethod exploits the remarkable zero-shot generation, code-writing, and in-context improvement capabilities of state-of-the-art LLMs, such as GPT-4, to perform evolutionary optimization over reward code. The resulting rewards can then be used to acquire complex skills via reinforcement learning. Without any task-specific prompting or pre-defined reward templates, \ourmethod generates reward functions that outperform expert human-engineered rewards. In a diverse suite of 29 open-source RL environments that include 10 distinct robot morphologies, \ourmethod outperforms human experts on \textbf{83\%} of the tasks, leading to an average normalized improvement of \textbf{52\%}. The generality of \ourmethod also enables a new gradient-free in-context learning approach to reinforcement learning from human feedback (RLHF), readily incorporating human inputs to improve the quality and the safety of the generated rewards without model updating. Finally, using \ourmethod rewards in a curriculum learning setting, we demonstrate for the first time, a simulated Shadow Hand capable of performing pen spinning tricks, adeptly manipulating a pen in circles at rapid speed.

\end{abstract}

\begin{figure*}[b] %
  \centering
  \includegraphics[width=0.9\textwidth]{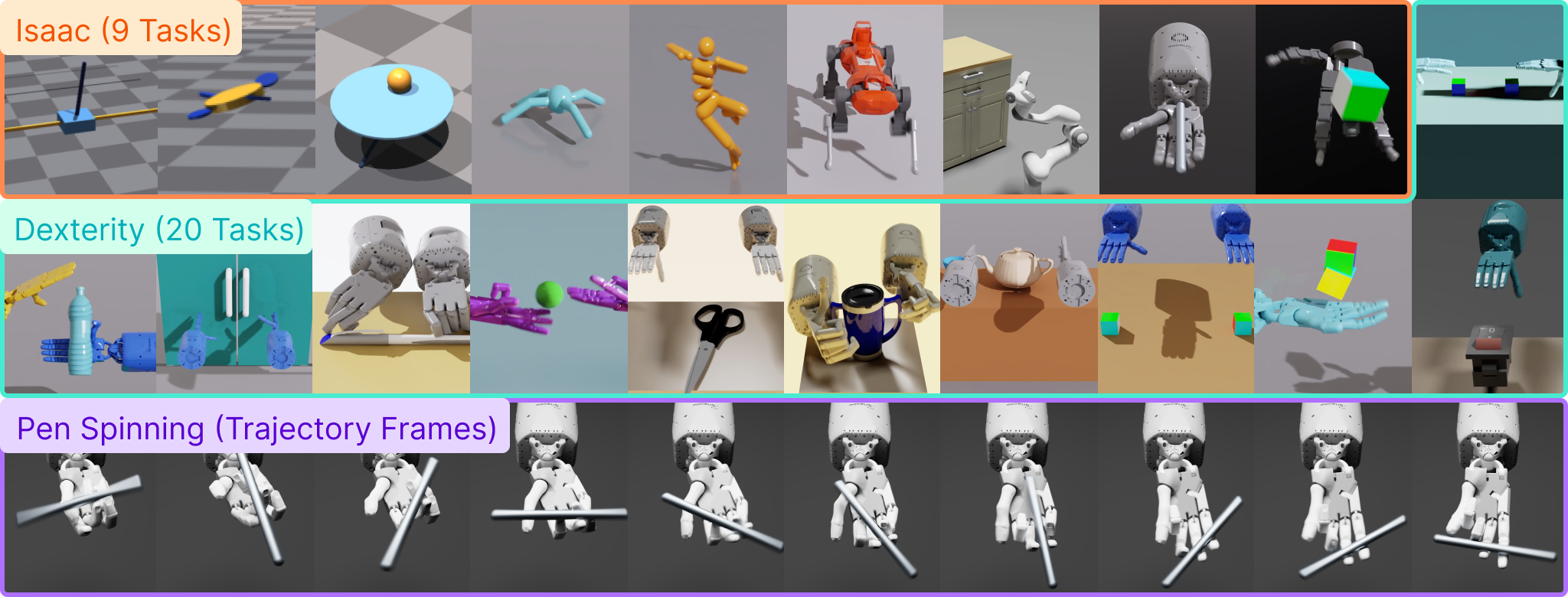} %
  \caption{\ourmethod generates human-level reward functions across diverse robots and tasks. Combined with curriculum learning, \ourmethod for the first time, unlocks rapid pen-spinning capabilities on an anthropomorphic five-finger hand.}
  \label{fig:eureka}
\end{figure*}

\section{Introduction}

Large Language Models (LLMs) have excelled as high-level semantic planners for robotics tasks~\citep{ahn2022can, singh2023progprompt}, but whether they can be used to learn complex low-level manipulation tasks, such as dexterous pen spinning, remains an open problem. Existing attempts require substantial domain expertise to construct task prompts or learn only simple skills, leaving a substantial gap in achieving human-level dexterity~\citep{yu2023language, brohan2023rt}. 

On the other hand, reinforcement learning (RL) has achieved impressive results in dexterity~\citep{andrychowicz2020learning, handa2023dextreme} as well as many other domains-if the human designers can carefully construct reward functions that accurately codify and provide learning signals for the desired behavior; likewise, many real-world RL tasks admit sparse rewards that are difficult for learning, necessitating reward shaping that provides incremental learning signals. Despite their fundamental importance, reward functions are known to be notoriously difficult to design in practice~\citep{russell1995artificial, sutton2018reinforcement}; a recent survey conducted finds 92\% of polled reinforcement learning researchers and practitioners report manual trial-and-error reward design and 89\% indicate that their designed rewards are sub-optimal~\citep{booth2023perils} and lead to unintended behavior~\citep{hadfield2017inverse}.

Given the paramount importance of reward design, we ask whether it is possible to develop a \textit{universal} reward programming algorithm using state-of-the-art coding LLMs, such as GPT-4. Their remarkable abilities in code writing, zero-shot generation, and in-context learning have previously enabled effective programmatic agents~\citep{shinn2023reflexion, wang2023voyager}. Ideally, this reward design algorithm should achieve human-level reward generation capabilities that scale to a broad spectrum of tasks, including dexterity, automate the tedious trial-and-error procedure without human supervision, and yet be compatible with human oversight to assure safety and alignment.

\begin{figure*}[t!]
\centering 
\includegraphics[width=\textwidth]{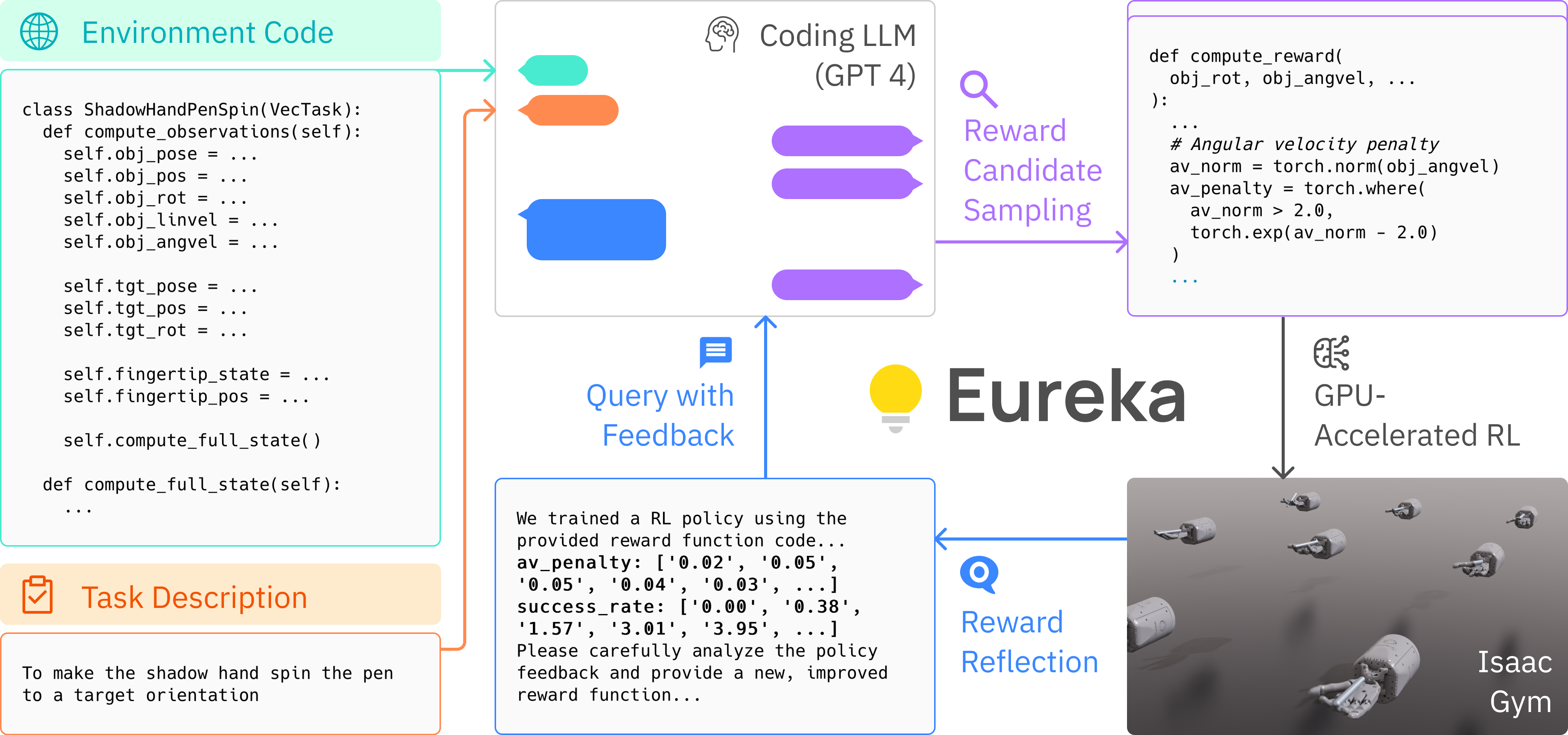}
\caption{\ourmethod takes unmodified environment source code and language task description as context to zero-shot generate executable reward functions from a coding LLM. Then, it iterates between reward sampling, GPU-accelerated reward evaluation, and reward reflection to progressively improve its reward outputs.}
  \label{fig:eureka-concept}
\end{figure*}

We introduce \textbf{E}volution-driven \textbf{U}niversal \textbf{RE}ward \textbf{K}it for \textbf{A}gent (\ourmethod), a novel reward design algorithm powered by coding LLMs with the following contributions:
\begin{enumerate}[leftmargin=*]
\item \textbf{Achieves human-level performance on reward design} across a diverse suite of 29 open-sourced RL environments that include 10 distinct robot morphologies, including quadruped, quadcopter, biped, manipulator, as well as several dexterous hands; see Fig.~\ref{fig:eureka}. Without any task-specific prompting or reward templates, \ourmethod autonomously generates rewards that outperform expert human rewards on \textbf{83\%} of the tasks and realizes an average normalized improvement of \textbf{52\%}.

\item \textbf{Solves dexterous manipulation tasks that were previously not feasible by manual reward engineering}. We consider pen spinning, in which a five-finger hand needs to rapidly rotate a pen in pre-defined spinning configurations for as many cycles as possible. Combining \ourmethod with curriculum learning, we demonstrate for the first time rapid pen spinning maneuvers on a simulated anthropomorphic Shadow Hand (see Figure~\ref{fig:eureka} bottom). 

\item \textbf{Enables a new \textit{gradient-free} in-context learning approach to reinforcement learning from human feedback (RLHF)} that can generate more performant and human-aligned reward functions based on various forms of human inputs without model updating. We demonstrate that \ourmethod can readily benefit from and improve upon existing human reward functions. Likewise, we showcase \ourmethod's capability in using purely textual feedback to generate progressively more human-aligned reward functions.

\end{enumerate}

Unlike prior work L2R on using LLMs to aid reward design~\citep{yu2023language}, \ourmethod is completely free of task-specific prompts, reward templates, as well as few-shot examples. In our experiments, \ourmethod significantly outperforms L2R due to its ability to generate free-form, expressive reward programs. \ourmethod's generality is made possible through three key algorithmic design choices: environment as context, evolutionary search, and reward reflection. First, by taking the \textbf{environment source code as context}, \ourmethod can zero-shot generate executable reward functions from the backbone coding LLM (GPT-4). Then, \ourmethod substantially improves the quality of its rewards by performing \textbf{evolutionary search}, iteratively proposing batches of reward candidates and refining the most promising ones within the LLM context window. This in-context improvement is made effective via \textbf{reward reflection}, a textual summary of the reward quality based on policy training statistics that enables automated and targeted reward editing; see Fig.~\ref{fig:eureka-reward} for an example of \ourmethod zero-shot reward as well as various improvements accumulated during its optimization. To ensure that \ourmethod can scale up its reward search to maximum potential, \ourmethod evaluates intermediate rewards using GPU-accelerated distributed reinforcement learning on IsaacGym~\citep{makoviychuk2021isaac}, which offers up to three orders of magnitude in policy learning speed, making \ourmethod an extensive algorithm that scales naturally with more compute. See Fig.~\ref{fig:eureka-concept} for an overview. We are committed to open-sourcing all prompts, environments, and generated reward functions to promote further research on LLM-based reward design.

\section{Problem Setting and Definitions}
The goal of reward design is to return a shaped reward function for a ground-truth reward function that may be difficult to optimize directly (e.g., sparse rewards); this ground-truth reward function may only be accessed via queries by the designer.  We first introduce the formal definition from~\citet{singh2010where}, which we then adapt to the program synthesis setting, which we call \textit{reward generation}. 

\begin{definition} 
\label{def:rdp}
(Reward Design Problem~\citep{singh2010where}) A \emph{reward design problem} (RDP) is a tuple $P = \langle M, \mathcal{R}, \pi_M, F \rangle$, where $M = (S,A,T)$ is the \textit{world model} with state space $S$, action space $A$, and transition function $T$. $\mathcal{R}$ is the space of reward functions; $\mathcal{A}_M(\cdot): \mathcal{R} \rightarrow \Pi$ is a learning algorithm that outputs a policy $\pi:S \rightarrow \Delta(A)$ that optimizes reward $R \in \mathcal{R}$ in the resulting \textit{Markov Decision Process} (MDP), $(M,R)$; $F: \Pi \rightarrow \mathbb{R}$ is the \textit{fitness} function that produces a scalar evaluation of any policy, which may only be accessed via policy queries (i.e., evaluate the policy using the ground truth reward function). In an RDP, the goal is to output a reward function $R \in \mathcal{R}$ 
such that the policy $\pi:=\mathcal{A}_M(R)$ that optimizes $R$ achieves the highest fitness score $F(\pi)$.
\end{definition}

\textbf{Reward Generation Problem.} In our problem setting, every component within a RDP is specified via code. 
Then, given a string $l$ that specifies the task, the objective of the reward generation problem is to output a reward function code $R$ such that $F(\mathcal{A}_{M}(R))$ is maximized.

\section{Method}
\ourmethod consists of three algorithmic components: 1) environment as context that enables zero-shot generation of executable rewards, 2) evolutionary search that iteratively proposes and refines reward candidates, and 3) reward reflection that enables fine-grained reward improvement. See Alg.~\ref{algo:eureka} for pseudocode; all prompts are included in App.~\ref{appendix:full-prompts}.

\subsection{Environment as Context}
\label{sec:env-as-context}

Reward design requires the environment specification to be provided to the LLM.
We propose directly feeding the raw environment source code (without the reward code, if exists) as context. \arxiv{Given that any reward function is a function over the environment's state and action variables, the only requirement in the source code is that it exposes these environment variables, which is easy to satisfy. In cases where the source code is not available, relevant state information can also be supplied via an API, for example. In practice, to ensure that the environment code fits within the LLM's context window and does not leak simulation internals (so that we can expect the same prompt to generalize to new simulators), we have an automatic script to extract just the environment code snippets that expose and fully specify the environment state and action variables. see App.~\ref{appendix:experimental-details} for details.}

\begin{wrapfigure}{r}{0.6\linewidth}
\begin{minipage}{0.6\textwidth}
\begin{algorithm}[H]
\caption{\ourmethod}
\label{algo:eureka}
\begin{algorithmic}[1]
\small
\STATE \textbf{Require}: Task description $l$, environment code $M$, \\ 
coding LLM $\texttt{LLM}$, fitness function $F$, initial prompt $\texttt{prompt}$
\STATE \textbf{Hyperparameters}: search iteration $N$, iteration batch size $K$
\FOR{\text{N iterations}}
\STATE \textcolor{purple}{\texttt{// Sample $K$ reward code from LLM}}
\STATE $\texttt{R}_1,..., \texttt{R}_k \sim \texttt{LLM}(l, M, \texttt{prompt})$ 
\STATE \textcolor{purple}{\texttt{// Evaluate reward candidates}}
\STATE $s_1 =F(\texttt{R}_1), ..., s_K = F(\texttt{R}_K)$
\STATE \textcolor{purple}{\texttt{// Reward reflection}}
\STATE $\texttt{prompt} := \texttt{prompt} : \texttt{Reflection}(R^n_{best}, s^n_{best})$,\\ 
where $best = \arg \max_{k} s_1,...,s_K$
\STATE \textcolor{purple}{\texttt{// Update Eureka reward}}
\STATE $R_{\text{Eureka}}, s_{\text{Eureka}} = (R^n_{\text{best}}, s^n_{\text{best}})$, \quad if $s^n_{\text{best}} > s_{\text{Eureka}}$
\ENDFOR
\STATE \textbf{Output}: $R_{\text{Eureka}}$
\end{algorithmic}
\end{algorithm}
\end{minipage}
\end{wrapfigure}

\begin{figure*}[t!]
\centering 
\includegraphics[width=0.8\textwidth]{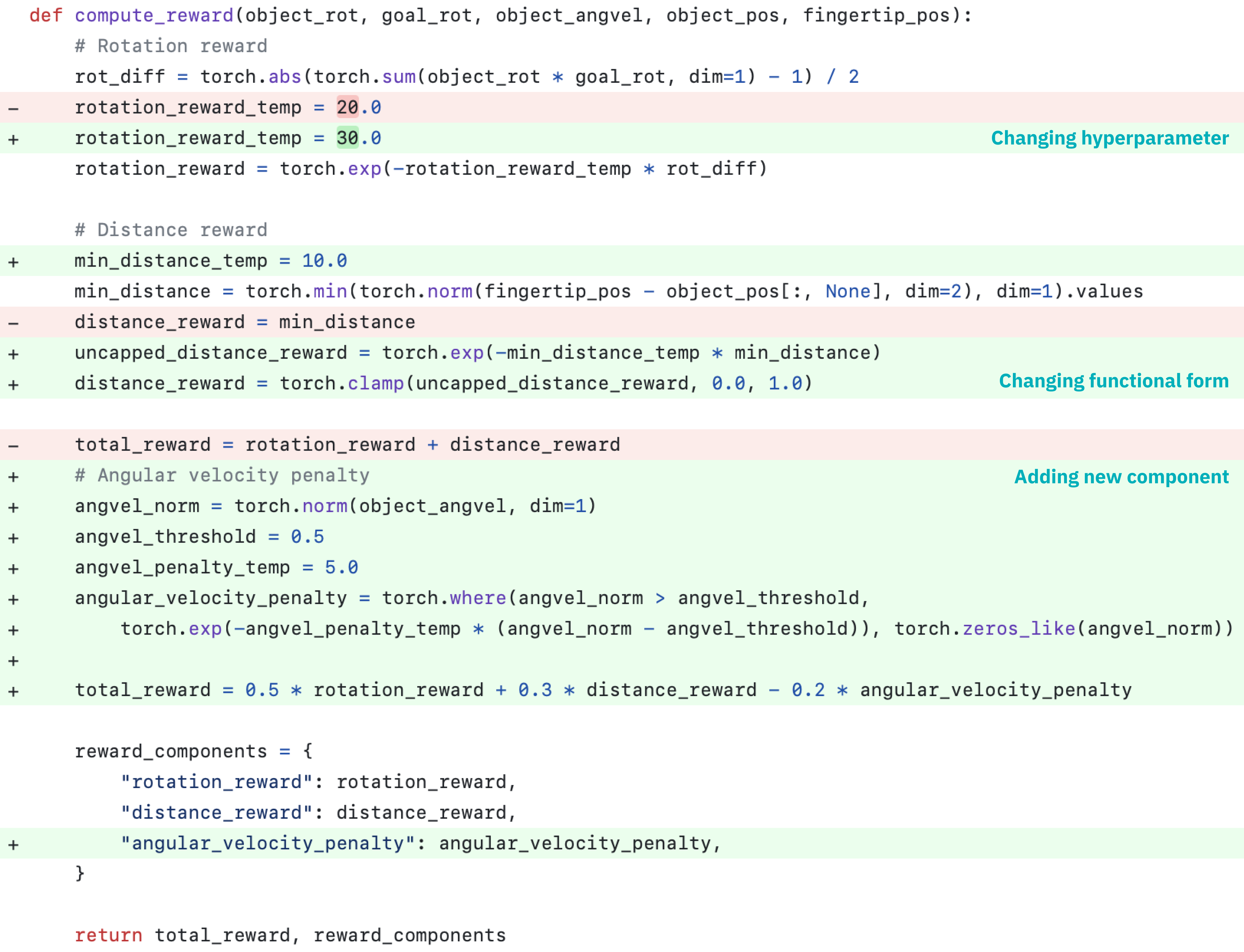}
\caption{\ourmethod can zero-shot generate executable rewards and then flexibly improve them with many distinct types of free-form modification, such as (1) changing the hyperparameter of existing reward components, (2) changing the functional form of existing reward components, and (3) introducing new reward components.}
  \label{fig:eureka-reward}
\end{figure*}

\arxiv{Given environment as context}, \ourmethod instructs the coding LLM to directly return executable Python code with only generic reward design and formatting tips, such as exposing individual components in the reward as a dictionary output (for reasons that will be apparent in Sec.~\ref{sec:reward-reflection}); \arxiv{see Prompt 1 and 3 in App.~\ref{appendix:full-prompts}.}
Remarkably, with only these minimal instructions, \ourmethod can already zero-shot generate plausibly-looking rewards in diverse environments in its first attempts. An example \ourmethod output is shown in Fig.~\ref{fig:eureka-reward}. As seen, \ourmethod adeptly composes over existing observation variables (e.g., \texttt{fingertip\_pos}) in the provided environment code and produces a competent reward code -- all without any environment-specific prompt engineering or reward templating. On the first try, however, the generated reward may not always be executable, and even if it is, it can be quite sub-optimal with respect to the task fitness metric $F$. While we can improve the prompt with task-specific formatting and reward design hints, doing so does not scale to new tasks and hinders the overall generality of our system. How can we effectively overcome the sub-optimality of single-sample reward generation?

\subsection{Evolutionary Search}
\label{sec:evolutionary-search}

In this section, we will demonstrate how evolutionary search presents a natural solution that addresses the aforementioned execution error and sub-optimality challenges. In each iteration, \ourmethod samples several independent outputs from the LLM (Line 5 in Alg.~\ref{algo:eureka}). Since the generations are i.i.d, the probability that \textit{all} reward functions from an iteration are buggy \textit{exponentially} decreases as the number of samples increases. We find that for all environments we consider, sampling just \arxiv{a modest number of samples (16)} contains at least one executable reward code in the first iteration.

Given executable reward functions from an earlier iteration, \ourmethod performs in-context \textit{reward mutation}, proposing new improved reward functions from the best one in the previous iteration.
\arxiv{Concretely, a new \ourmethod iteration will take the best-performing reward from the previous iteration, its reward reflection (Sec.~\ref{sec:reward-reflection}), and the mutation prompt (Prompt 2 in App.~\ref{appendix:full-prompts})} as context and generate $K$ more i.i.d reward outputs from the LLM; several illustrative reward modifications are visualized in Fig.~\ref{fig:eureka-reward}. This iterative optimization continues until a specified number of iterations has been reached. Finally, we perform multiple random restarts to find better maxima; this is a standard strategy in global optimization. In all our experiments, \ourmethod conducts 5 independent runs per environment, and for each run, searches for $5$ iterations with $K=16$ samples per iteration.

\subsection{Reward Reflection}
\label{sec:reward-reflection}
In order to ground the in-context reward mutation, we must be able to put into words the quality of the generated rewards. We propose \arxiv{\textit{reward reflection}, an automated feedback that summarizes the policy training dynamics in texts. Specifically, given that \ourmethod reward functions are asked to expose their individual components in the reward program (e.g., \texttt{reward\_components} in Fig.~\ref{fig:eureka-reward}), reward reflection tracks the scalar values of all reward components and the task fitness function at intermediate policy checkpoints throughout training. For instance, consider the illustrative example in Fig.~\ref{fig:eureka-concept}, where the snapshot values of $\texttt{av\_penalty}$ are provided as a list in the reward feedback. See App.~\ref{appendix:reward-reflection-examples} for full examples.}

This reward reflection procedure, though simple to construct, is important due to \arxiv{two reasons: (1) the lack of fine-grained reward improvement signal in the task fitness function, and (2) the algorithm-dependent nature of reward optimization~\citep{booth2023perils}.} First, as we can query the task fitness function $F$ on the resulting policies, a simple strategy is to just provide this numerical score as the reward evaluation. While serving as the holistic ground-truth metric, the task fitness function itself lacks in credit assignment, providing no useful information on \textit{why} a reward function works or not. Second, whether a reward function is effective is influenced by the particular choice of RL algorithm, and the same reward may perform very differently even under the same optimizer given hyperparameter differences~\citep{henderson2018deep, agarwal2021deep}. By providing detailed accounts on how well the RL algorithm optimizes individual reward components, reward reflection enables \ourmethod to produce more intricate and targeted reward editing.

\section{Experiments}

We thoroughly evaluate \ourmethod on a diverse suite of robot embodiments and tasks, testing its ability to generate reward functions, solve new tasks, and incorporate various forms of human input. We use GPT-4~\citep{openai2023gpt4}, in particular the \texttt{gpt-4-0314} variant, as the backbone LLM for all LLM-based reward-design algorithms unless specified otherwise. 

\textbf{Environments.} Our environments consist of 10 distinct robots and 29 tasks implemented using the IsaacGym simulator~\citep{makoviychuk2021isaac}. First, we include 9 original environments from IsaacGym (\textbf{Isaac}), covering a diverse set of robot morphologies from quadruped, bipedal, quadrotor, cobot arm, to dexterous hands. In addition to coverage over robot form factors, we ensure \textit{depth} in our evaluation by including all 20 tasks from the Bidexterous Manipulation (\textbf{Dexterity}) benchmark~\citep{chen2022towards}. Dexterity contains 20 complex bi-manual tasks that require a pair of Shadow Hands to solve a wide range of complex manipulation skills, ranging from object handover to rotating a cup by 180 degrees. For the task description input to \ourmethod, we use the official description provided in the environment repository when possible. See App.~\ref{appendix:environment-details} for details on all environments. It is worth noting that both benchmarks are publicly released concurrently, or after the GPT-4 knowledge cut-off date (September 2021), so GPT-4 is unlikely to have accumulated extensive internet knowledge about these tasks, making them ideal testbeds for assessing \ourmethod's reward generation capability compared to measurable human-engineered reward functions.

\begin{figure*}[t!]
\centering 
\includegraphics[width=\textwidth]{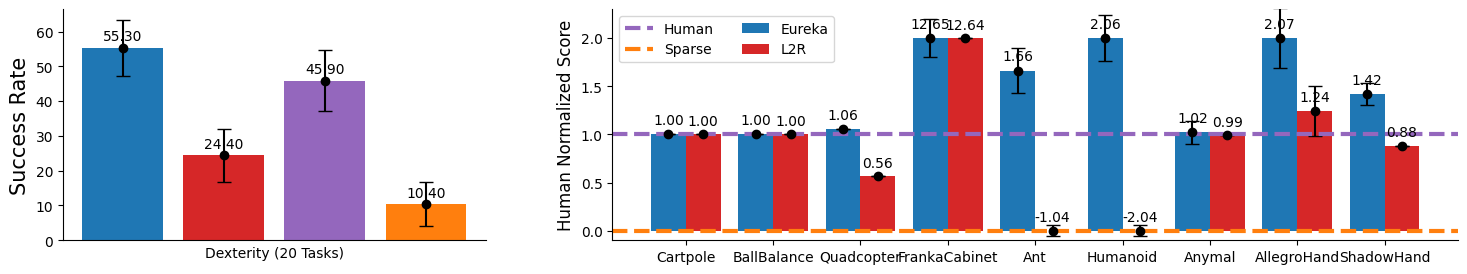}
\caption{\ourmethod outperforms Human and L2R across all tasks. In particular, \ourmethod realizes much greater gains on high-dimensional dexterity environments.}
  \label{fig:eureka-bar}
\end{figure*}

\subsection{Baselines}

\textbf{L2R}~\citep{yu2023language} proposes a two-stage LLM-prompting solution to generate templated rewards. For an environment and task specified in natural language, a first LLM is asked to fill in a natural language template describing the agent's motion; then, a second LLM is asked to convert this ``motion description'' into code that calls a manually defined set of reward API primitives to write a reward program that sets their parameters. To make L2R competitive for our tasks, we define the motion description template to mimic the original L2R templates, and we construct the API reward primitives using the individual components of the original human rewards when possible. Note that this gives L2R an advantage as it has access to the original reward functions. Consistent with \ourmethod, we conduct 5 independent L2R runs per environment, and for each run, we generate 16 reward samples. See App.~\ref{appendix:baseline-details} for more details.

\textbf{Human.} These are the original shaped reward functions provided in our benchmark tasks. As these reward functions are written by active reinforcement learning researchers who designed the tasks, these reward functions represent the outcomes of expert-level human reward engineering.

\textbf{Sparse.} These are identical to the fitness functions $F$ that we use to evaluate the quality of the generated rewards. Like Human, these are also provided by the benchmark. On the dexterity tasks, they are uniformly binary indicator functions that measure task success; on Isaac tasks, they vary in functional forms depending on the nature of the task. See App.~\ref{appendix:environment-details} for a description of the ground-truth scoring metric for all tasks. 

\subsection{Training Details}
\textbf{Policy Learning.} For each task, all final reward functions are optimized using the same RL algorithm with the same set of hyperparameters. Isaac and Dexterity share a well-tuned PPO implementation~\citep{schulman2017proximal, rl-games2021}, and we use this implementation and the task-specific PPO hyperparameters without any modification. Note that these task hyperparameters are tuned to make the official human-engineered rewards work well. \arxiv{For each final reward function obtained from each method,} we run 5 independent PPO training runs and report the average of the maximum task metric values achieved \arxiv{from 10 policy checkpoints sampled at fixed intervals}. \arxiv{In particular, the maximum is taken over the same number of checkpoints for each approach.}

\textbf{Reward Evaluation Metrics.} For Isaac tasks, since the task metric $F$ for each task varies in semantic meaning and scale, we report the \textbf{human normalized score} for \ourmethod and L2R, $\frac{\texttt{Method} - \texttt{Sparse}}{\lvert \texttt{Human} - \texttt{Sparse} \rvert}$. This metric provides a holistic measure of how \ourmethod rewards fare against human-expert rewards with respect to the ground-truth task metric. For Dexterity, since all tasks are evaluated using the binary success function, we directly report success rates.

\subsection{Results}
\textbf{\ourmethod outperforms human rewards.} In Figure~\ref{fig:eureka-bar}, we report the aggregate results on Dexterity and Isaac. Notably, \ourmethod exceeds or performs on par to human level on \textit{all} Isaac tasks and 15 out of 20 tasks on Dexterity (see App.~\ref{appendix:additional-results} for a per-task breakdown). In contrast, L2R, while comparable on low-dimensional tasks (e.g., CartPole, BallBalance), lags significantly behind on high-dimensional tasks. Despite being provided access to some of the same reward components as Human, L2R still underperforms \ourmethod after its initial iteration, when both methods have had the same number of reward queries. As expected, L2R's lack of expressivity severely limits its performance. In contrast, \ourmethod generates free-form rewards from scratch without any domain-specific knowledge and performs substantially better. \arxiv{In App.~\ref{appendix:additional-results}, we present results on additional evaluation metrics such as interquantile mean (IQM), probability of improvement~\citep{agarwal2021deep}, and the aggregate RL training curves; on all evaluations, we observe the consistent trend that \ourmethod generates the most capable reward functions.} Furthermore, we ablate GPT-4 with GPT-3.5 and find \ourmethod degrades in performance but still matches or exceeds human-level on most Isaac tasks, indicating that its general principles can be readily applied to coding LLMs of varying qualities.

\begin{wrapfigure}{r}{0.6\textwidth}
\centering 
\includegraphics[width=0.6\textwidth]{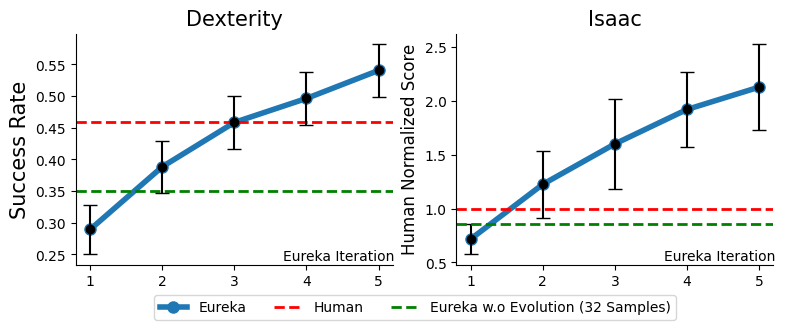}
\caption{\ourmethod progressively produces better rewards via in-context evolutionary reward search.}
\vspace{-0.3cm}
  \label{fig:eureka-improvement}
\end{wrapfigure}

\textbf{\ourmethod consistently improves over time.} 
In Fig.~\ref{fig:eureka-improvement}, we 
visualize the average performance of the cumulative best \ourmethod rewards after each evolution iteration. Moreover, we study an ablation, \textbf{\ourmethod w.o. Evolution (32 Samples)}, which performs only the initial reward generation step, sampling the same number of reward functions as two iterations in the original \ourmethod. This ablation helps study, given a fixed number of reward function budget, whether it is more advantageous to perform the \ourmethod evolution or simply sample more first-attempt rewards without iterative improvement. As seen, on both benchmarks, \ourmethod rewards steadily improve and eventually surpass human rewards in performance despite sub-par initial performances. This consistent improvement also cannot be replaced by just sampling more in the first iteration as the ablation's performances are lower than \ourmethod after 2 iterations on both benchmarks. Together, these results demonstrate that \ourmethod's novel evolutionary optimization is indispensable for its final performance.

\begin{wrapfigure}{r}{0.4\textwidth}
\centering 
\includegraphics[width=0.4\textwidth]{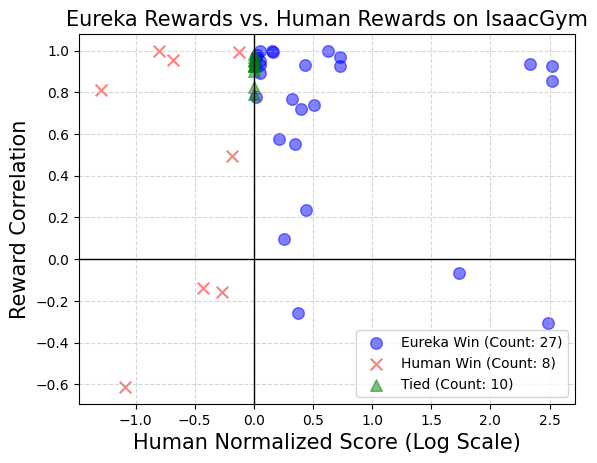}
\vspace{-0.5cm}
\caption{Eureka generates novel rewards.}
  \label{fig:eureka-correlation}
\end{wrapfigure}

\textbf{\ourmethod generates novel rewards.} 
We assess the novelty of \ourmethod rewards by computing the \textit{correlations} between \ourmethod and human rewards on all the Isaac tasks; see App.~\ref{appendix:environment-details} for details on this procedure. Then, we plot the correlations against the human normalized scores on a scatter-plot in Figure~\ref{fig:eureka-correlation}, where each point represents a single \ourmethod reward on a single task. As shown, \ourmethod mostly generates weakly correlated reward functions that outperform the human ones. In addition, by examining the average correlation by task (App.~\ref{appendix:additional-results}), we observe that \textit{the harder the task is, the less correlated the \ourmethod rewards}. We hypothesize that human rewards are less likely to be near optimal for difficult tasks, leaving more room for \ourmethod rewards to be different and better. In a few cases, \ourmethod rewards are even \textit{negatively} correlated with human rewards but perform significantly better, demonstrating that \ourmethod can \textit{discover} novel reward design principles that may run counter to human intuition; we illustrate these \ourmethod rewards in App.~\ref{appendix:negative-correlated-examples}.

\textbf{Reward reflection enables targeted improvement.} 
To assess the importance of constructing reward reflection in the reward feedback, we evaluate an ablation, \textbf{\ourmethod (No Reward Reflection)}, which reduces the reward feedback prompt to include only snapshot values of the task metric $F$. Averaged over all Isaac tasks, \ourmethod without reward reflection reduces the average normalized score by 28.6\%; in App.~\ref{appendix:additional-results}, we provide detailed per-task breakdown and observe much greater performance deterioration on higher dimensional tasks. To provide qualitative analysis, in App.~\ref{appendix:reward-reflection-examples}, we include several examples in which \ourmethod utilizes the reward reflection to perform targeted reward editing.

\begin{wrapfigure}{r}{0.4\textwidth}
\centering 
\includegraphics[width=0.4\textwidth]{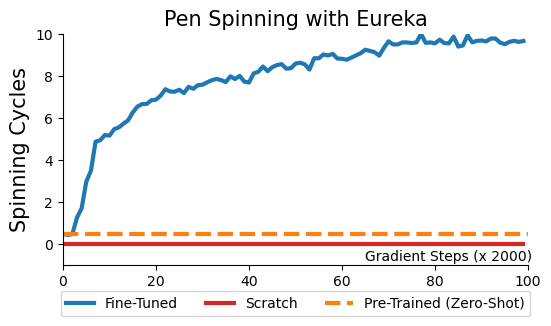}
\caption{\ourmethod can be flexibly combined with curriculum learning to acquire complex dexterous skills.}
\vspace{-0.5cm}
  \label{fig:eureka-pen-spinning}
\end{wrapfigure}

\textbf{\ourmethod with curriculum learning enables dexterous pen spinning.} 
Finally, we investigate whether \ourmethod can be used to solve a truly novel and challenging dexterous task. To this end, we propose pen spinning as a test bed. This task is highly dynamic and requires a Shadow Hand to continuously rotate a pen to achieve some pre-defined spinning patterns for as many cycles as possible; \arxiv{we implement this task on top of the original Shadow Hand environment in Isaac Gym without changes to any physics parameter, ensuring physical realism.} We consider a \textit{curriculum learning}~\citep{bengio2009curriculum} approach to break down the task into manageable components that can be independently solved by \ourmethod. Specifically, we first use \ourmethod to generate a reward for \arxiv{the task of} re-orienting the pen to random target configurations \arxiv{and train a policy using the final \ourmethod reward}. Then, using this pre-trained policy (\textbf{Pre-Trained}), we fine-tune it using the \arxiv{same} \ourmethod reward to reach the sequence of pen-spinning configurations (\textbf{Fine-Tuned}). To demonstrate the importance of curriculum learning, we also directly train a policy from scratch \arxiv{on the target task} using \ourmethod reward without the first-stage pre-training (\textbf{Scratch}). The RL training curves are shown in Figure~\ref{fig:eureka-pen-spinning}. Eureka fine-tuning quickly adapts the policy to successfully spin the pen for many cycles in a row; see project website for videos. In contrast, neither pre-trained or learning-from-scratch policies can complete even a single cycle of pen spinning. In addition, using this \ourmethod fine-tuning approach, we have also trained pen spinning policies for a variety of different spinning configurations; all pen spinning videos can be viewed on our project website, and experimental details are in App.~\ref{appendix:eureka-pen-spinning}. These results demonstrate \ourmethod's applicability to advanced policy learning approaches, which are often necessary for learning very complex skills

\subsection{\ourmethod From Human Feedback}
\label{sec:eureka-human-feedback}
In addition to automated reward design, \ourmethod enables a new gradient-free in-context learning approach to RL from Human Feedback (RLHF) that can readily incorporate various types of human inputs to generate more performant and human-aligned reward functions.

\begin{wrapfigure}{r}{0.5\textwidth}
\centering 
\includegraphics[width=0.5\textwidth]{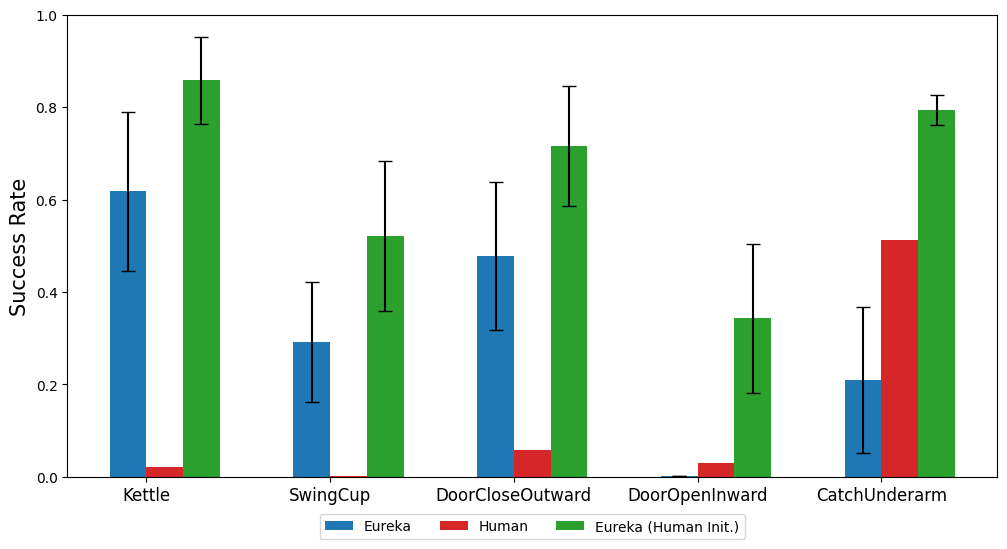}
\caption{\small{\ourmethod effectively improves and benefits from human reward initialization.}}
  \label{fig:eureka-initialization}
\end{wrapfigure}

\textbf{\ourmethod can improve and benefit from human reward functions.} We study whether starting with a human reward function initialization, a common scenario in real-world RL applications, is advantageous for \ourmethod. Importantly, incorporating human initialization requires no modification to \ourmethod\ -- we can simply substitute the raw human reward function as the output of the first \ourmethod iteration. To investigate this, we select several tasks from Dexterity that differ in the relative performances between the original \ourmethod and human rewards. The full results are shown in Figure~\ref{fig:eureka-initialization}.

As shown, regardless of the quality of the human rewards, \ourmethod improves and benefits from human rewards as \textbf{\ourmethod (Human Init.)} is uniformly better than both \ourmethod and Human on all tasks. This suggests that \ourmethod's in-context reward improvement capability is largely independent of the quality of the base reward. Furthermore, the fact that \ourmethod can significantly improve over human rewards even when they are highly sub-optimal hints towards an interesting hypothesis: \textit{human designers are generally knowledgeable about relevant state variables but are less proficient at designing rewards using them}. This makes intuitive sense as identifying relevant state variables that should be included in the reward function involves mostly common sense reasoning, but reward design requires specialized knowledge and experience in RL. Together, these results demonstrate \ourmethod's \textit{reward assistant} capability, perfectly complementing human designers' knowledge about useful state variables and making up for their less proficiency on how to design rewards using them. In App.~\ref{appendix:human-initialization-examples}, we provide several examples of \ourmethod (Human Init.) steps.

\begin{wrapfigure}{r}{0.5\textwidth}
\centering 
\includegraphics[width=0.5\textwidth]{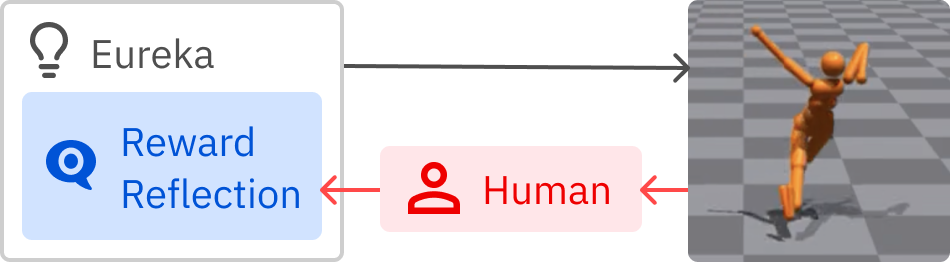}
\vspace{0.2cm}
\resizebox{0.5\textwidth}{!}{
\begin{tabular}[c]{l|cc}\toprule
Method & Forward Velocity
& Human Preference \\ \midrule
\ourmethod & \textbf{7.53} & 5/20\\
\ourmethod-HF & 5.58 & \textbf{15/20}\\ 
\bottomrule
\end{tabular}}
\vspace{-0.4cm}
\caption{\ourmethod can incorporate human reward reflection to modify rewards that induce safer and more human-aligned behavior.}
  \label{fig:eureka-rlhf}
\end{wrapfigure}

\textbf{Reward reflection via human feedback induces aligned behavior.} So far, all \ourmethod rewards are optimized against a fixed, black-box task fitness function $F$. This task metric, however, may not fully align with human intent. Moreover, in many open-ended tasks, $F$ may not be available in the first place~\citep{fan2022minedojo}. In these challenging scenarios, we propose to augment \ourmethod by having humans step in and put into words the reward reflection in terms of the desired behavior and correction. We investigate this capability in \ourmethod by teaching a Humanoid agent how to run purely from textual reward reflection; in App.~\ref{appendix:human-reward-reflection-example}, we show the exact sequence of human feedback and \ourmethod rewards. Then, we conduct a user study asking 20 unfamiliar users to indicate their preferences between two policy rollout videos shown in random order, one trained with human reward reflection (\textbf{\ourmethod-HF}) and the other one trained with the original best \ourmethod reward; the details are in App.~\ref{appendix:eureka-rlhf-details}. As shown in Fig.~\ref{fig:eureka-rlhf}, despite running a bit slower, the \ourmethod-HF agent is preferred by a large majority of our users. Qualitative, we indeed see that the \ourmethod-HF agent acquires safer and more stable gait, as instructed by the human. See the project website for a comparison.

\vspace{-0.3cm}
\section{Related Work}
\vspace{-0.3cm}

\textbf{Reward Design.} 
Reward engineering is a long-standing challenge in reinforcement learning~\citep{singh2010where, sutton2018reinforcement}. The most common reward design method is manual trial-and-error~\citep{knox2023reward, booth2023perils}. Inverse reinforcement learning (IRL) infers reward functions from demonstrations~\citep{abbeel2004apprenticeship, ziebart2008maximum, ho2016generative}, but it requires expensive expert data collection, which may not be available, and outputs non-interpretable black-box reward functions. Several prior works have studied automated reward search through evolutionary algorithms~\citep{niekum2010genetic, chiang2019learning, faust2019evolving}. These early attempts are limited to task-specific implementations of evolutionary algorithms that search over only parameters within provided reward templates. Recent works have also proposed using pretrained foundation models that can produce reward functions for new tasks~\citep{ma2022vip, ma2023liv, fan2022minedojo, du2023vision, karamcheti2023language, du2023guiding, kwon2023reward}. Most of these approaches output \textit{scalar} rewards that lack interpretability and do not naturally admit the capability to improve or adapt rewards on-the-fly. In contrast, \ourmethod adeptly generates free-form, white-box reward code and effectively in-context improves.

\textbf{Code Large Language Models for Decision Making.}
Recent works have considered using coding LLMs~\citep{austin2021program, chen2021evaluating, roziere2023code} to generate grounded and structured programmatic output for decision making and robotics problems~\citep{liang2023code, singh2023progprompt, wang2023demo2code, huang2023instruct2act, wang2023voyager, liu2023llm+, silver2023generalized, ding2023task, lin2023text2motion, xie2023translating}. However, most of these works rely on known motion primitives to carry out robot actions and do not apply to robot tasks that require low-level skill learning, such as dexterous manipulation. The closest to our work is a recent work~\citep{yu2023language} that also explores using LLMs to aid reward design. Their approach, however, requires domain-specific task descriptions and reward templates.

\vspace{-0.3cm}
\section{Conclusion}
\vspace{-0.3cm}

We have presented \ourmethod, a universal reward design algorithm powered by coding large language models and in-context evolutionary search. Without any task-specific prompt engineering or human intervention, \ourmethod achieves human-level reward generation on a wide range of robots and tasks. \ourmethod's particular strength in learning dexterity solves dexterous pen spinning for the first time with a curriculum learning approach. Finally, \ourmethod enables a gradient-free approach to reinforcement learning from human feedback that readily incorporates human reward initialization and textual feedback to better steer its reward generation. The versatility and substantial performance gains of \ourmethod suggest that the simple principle of combining large language models with evolutionary algorithms are a general and scalable approach to reward design, an insight that may be generally applicable to difficult, open-ended search problems.

\newpage 

\section*{Acknowledgement}
We are grateful to colleagues and friends at NVIDIA and UPenn for their helpful feedback and insightful discussions. We thank Viktor Makoviychuk, Yashraj Narang, Iretiayo Akinola, Erwin Coumans for their assistance on Isaac Gym experiment and rendering. This work is done during Yecheng Jason Ma's internship at NVIDIA. We acknowledge funding support from NSF CAREER Award 2239301, ONR award N00014-22-1-2677, NSF Award CCF-1917852, and ARO Award W911NF-20-1-0080.

\newpage 

\bibliography{iclr2024_conference}
\bibliographystyle{iclr2024_conference}

\newpage 

\appendix
\appendix
\addcontentsline{toc}{section}{Appendix} %
\part{Appendix} %
\parttoc %

\newpage 
\renewcommand\lstlistingname{Prompt}
\setcounter{lstlisting}{0}

\section{Full Prompts}
\label{appendix:full-prompts}
In this section, we provide all \ourmethod prompts. 

\begin{lstlisting}[basicstyle=\fontfamily{\ttdefault}\scriptsize, breaklines=true,caption={Initial system prompt}]
You are a reward engineer trying to write reward functions to solve reinforcement learning tasks as effective as possible.
Your goal is to write a reward function for the environment that will help the agent learn the task described in text. 
Your reward function should use useful variables from the environment as inputs. As an example,
the reward function signature can be: 
@torch.jit.script
def compute_reward(object_pos: torch.Tensor, goal_pos: torch.Tensor) -> Tuple[torch.Tensor, Dict[str, torch.Tensor]]:
    ...
    return reward, {}
Since the reward function will be decorated with @torch.jit.script,
please make sure that the code is compatible with TorchScript (e.g., use torch tensor instead of numpy array). 
Make sure any new tensor or variable you introduce is on the same device as the input tensors. 
\end{lstlisting}
\begin{lstlisting}[basicstyle=\fontfamily{\ttdefault}\scriptsize, breaklines=true,caption={Reward reflection and feedback}]
We trained a RL policy using the provided reward function code and tracked the values of the individual components in the reward function as well as global policy metrics such as success rates and episode lengths after every {epoch_freq} epochs and the maximum, mean, minimum values encountered:
<REWARD REFLECTION HERE>

Please carefully analyze the policy feedback and provide a new, improved reward function that can better solve the task. Some helpful tips for analyzing the policy feedback:
    (1) If the success rates are always near zero, then you must rewrite the entire reward function
    (2) If the values for a certain reward component are near identical throughout, then this means RL is not able to optimize this component as it is written. You may consider
        (a) Changing its scale or the value of its temperature parameter
        (b) Re-writing the reward component 
        (c) Discarding the reward component
    (3) If some reward components' magnitude is significantly larger, then you must re-scale its value to a proper range
Please analyze each existing reward component in the suggested manner above first, and then write the reward function code. 
\end{lstlisting}
\begin{lstlisting}[basicstyle=\fontfamily{\ttdefault}\scriptsize, breaklines=true,caption={Code formatting tip}]
The output of the reward function should consist of two items:
    (1) the total reward,
    (2) a dictionary of each individual reward component.
The code output should be formatted as a python code string: "```python ... ```".

Some helpful tips for writing the reward function code:
    (1) You may find it helpful to normalize the reward to a fixed range by applying transformations like torch.exp to the overall reward or its components
    (2) If you choose to transform a reward component, then you must also introduce a temperature parameter inside the transformation function; this parameter must be a named variable in the reward function and it must not be an input variable. Each transformed reward component should have its own temperature variable
    (3) Make sure the type of each input variable is correctly specified; a float input variable should not be specified as torch.Tensor
    (4) Most importantly, the reward code's input variables must contain only attributes of the provided environment class definition (namely, variables that have prefix self.). Under no circumstance can you introduce new input variables.
\end{lstlisting}

\section{Environment Details}
\label{appendix:environment-details}

In this section, we provide environment details. For each environment, we list its observation and action dimensions, the verbatim task description, and the task fitness function $F$. $F$ is evaluated per-environment, and our policy feedback uses the mean across environment instances.

For the functions below, \verb+||+ denotes the $L_2$ norm, and \verb|1[]| denotes the indicator function.

\begin{tabularx}{\textwidth}{X}
    \toprule
    \multicolumn{1}{c}{IsaacGym Environments} \\[1px]
    \hline
    Environment (obs dim, action dim) \\ Task description \\ Task fitness function $F$ \\[1px]
    \hline
    Cartpole (4, 1) \\ To balance a pole on a cart so that the pole stays upright \\ \verb|duration| \\ [1px]
    \hline
    Quadcopter (21, 12) \\ To make the quadcopter reach and hover near a fixed position \\ \verb|-cur_dist| \\[1px]
    \hline
    FrankaCabinet (23, 9) \\ To open the cabinet door \\ \verb|1[cabinet_pos > 0.39]| \\[1px]
    \hline
    Anymal (48, 12) \\ To make the quadruped follow randomly chosen x, y, and yaw target velocities \\ \verb|-(linvel_error +  angvel_error)| \\[1px]
    \hline
    BallBalance (48, 12) \\ To keep the ball on the table top without falling \\ \verb|duration| \\[1px]
    \hline
    Ant (60, 8) \\ To make the ant run forward as fast as possible \\ \verb|cur_dist - prev_dist| \\[1px]
    \hline
    AllegroHand (88, 16) \\ To make the hand spin the object to a target orientation \\ \verb|number of consecutive successes where| \\ \verb|current success is 1[rot_dist < 0.1]| \\[1px]
    \hline
    Humanoid (108, 21) \\ To make the humanoid run as fast as possible \\ \verb|cur_dist - prev_dist| \\[1px]
    \hline
    ShadowHand (211, 20) \\ To make the shadow hand spin the object to a target orientation \\ \verb|number of consecutive successes where| \\ \verb|current success is 1[rot_dist < 0.1]| \\[1px]
    \bottomrule
\end{tabularx}
\begin{tabularx}{\textwidth}{X}
    \toprule
    \multicolumn{1}{c}{Dexterity Environments} \\[1px]
    \hline
    Environment (obs dim, action dim) \\ Task description \\ Task fitness function $F$ \\
    \hline
    Over (398, 40) \\ This class corresponds to the HandOver task. This environment consists of two shadow hands with palms facing up, opposite each other, and an object that needs to be passed. In the beginning, the object will fall randomly in the area of the shadow hand on the right side. Then the hand  holds the object and passes the object to the other hand. Note that the base of the hand is  fixed. More importantly, the hand which holds the object initially can not directly touch the  target, nor can it directly roll the object to the other hand, so the object must be thrown up and stays in the air in the process \\ \verb|1[dist < 0.03]| \\[1px]
    \hline
    DoorCloseInward (417, 52) \\ This class corresponds to the DoorCloseInward task. This environment require a closed door to be opened and the door can only be pushed outward or initially open inward. Both these two environments only need to do the push behavior, so it is relatively simple \\ \verb|1[door_handle_dist < 0.5]| \\[1px]
    \hline
    DoorCloseOutward (417, 52) \\ This class corresponds to the DoorCloseOutward task. This environment also require a closed door  to be opened and the door can only be pushed inward or initially open outward, but because they  can't complete the task by simply pushing, which need to catch the handle by hand and then open or close it, so it is relatively difficult \\ \verb|1[door_handle_dist < 0.5]| \\[1px]
    \hline
    DoorOpenInward (417, 52) \\ This class corresponds to the DoorOpenInward task. This environment also require a opened door  to be closed and the door can only be pushed inward or initially open outward, but because they  can't complete the task by simply pushing, which need to catch the handle by hand and then open or close it, so it is relatively difficult \\ \verb|1[door_handle_dist > 0.5]| \\[1px]
    \hline
    DoorOpenOutward (417, 52) \\ This class corresponds to the DoorOpenOutward task. This environment require a opened door  to be closed and the door can only be pushed outward or initially open inward. Both these two  environments only need to do the push behavior, so it is relatively simple \\ \verb|1[door_handle_dist < 0.5]| \\[1px]
    \hline
    Scissors (417, 52) \\ This class corresponds to the Scissors task. This environment involves two hands and scissors,  we need to use two hands to open the scissors \\ \verb|1[dof_pos > -0.3]| \\[1px]
    \hline
    SwingCup (417, 52) \\ This class corresponds to the SwingCup task. This environment involves two hands and a dual handle cup, we need to use two hands to hold and swing the cup together \\ \verb|1[rot_dist < 0.785]| \\[1px]
    \hline
    Switch (417, 52) \\ This class corresponds to the Switch task. This environment involves dual hands and a bottle, we need to use dual hand fingers to press the desired button \\ \verb|1[1.4 - (left_switch_z + right_switch_z) > 0.05]| \\[1px]
    \hline
    Kettle (417, 52) \\ This class corresponds to the PourWater task. This environment involves two hands, a kettle, and a bucket, we need to hold the kettle with one hand and the bucket with the other hand, and pour the water from the kettle into the bucket. In the practice task in Isaac Gym, we use many small balls to simulate the water \\ \verb+1[|bucket - kettle_spout| < 0.05]+ \\[1px]
    \hline
    LiftUnderarm (417, 52) \\ This class corresponds to the LiftUnderarm task. This environment requires grasping the pot handle  with two hands and lifting the pot to the designated position. This environment is designed to simulate the scene of lift in daily life and is a practical skill \\ \verb|1[dist < 0.05]| \\[1px]
    \hline
    Pen (417, 52) \\ This class corresponds to the Open Pen Cap task. This environment involves two hands and a pen, we need to use two hands to open the pen cap \\ \verb+1[5 * |pen_cap - pen_body| > 1.5]+ \\[1px]
    \hline
    BottleCap (420, 52) \\ This class corresponds to the Bottle Cap task. This environment involves two hands and a bottle, we  need to hold the bottle with one hand and open the bottle cap with the other hand. This skill requires  the cooperation of two hands to ensure that the cap does not fall \\ \verb|1[dist > 0.03]| \\[1px]
    \hline
    CatchAbreast (422, 52) \\ This class corresponds to the Catch Abreast task. This environment consists of two shadow hands placed side by side in the same direction and an object that needs to be passed. Compared with the previous  environment which is more like passing objects between the hands of two people, this environment is designed to simulate the two hands of the same person passing objects, so different catch techniques are also required and require more hand translation and rotation techniques \\ \verb|1[dist] < 0.03| \\[1px]
    \hline
    CatchOver2Underarm (422, 52) \\ This class corresponds to the Over2Underarm task. This environment is similar to Catch Underarm,  but with an object in each hand and the corresponding goal on the other hand. Therefore, the environment  requires two objects to be thrown into the other hand at the same time, which requires a higher  manipulation technique than the environment of a single object \\ \verb|1[dist < 0.03]| \\[1px]
    \hline
    CatchUnderarm (422, 52) \\ This class corresponds to the Catch Underarm task. In this task, two shadow hands with palms facing upwards are controlled to pass an object from one palm to the other. What makes it more difficult  than the Hand over task is that the hands' translation and rotation degrees of freedom are no longer frozen but are added into the action space \\ \verb|1[dist < 0.03]| \\[1px]
    \hline
    ReOrientation (422, 40) \\ This class corresponds to the ReOrientation task. This environment involves two hands and two objects.  Each hand holds an object and we need to reorient the object to the target orientation \\ \verb|1[rot_dist < 0.1]| \\[1px]
    \hline
    GraspAndPlace (425, 52) \\ This class corresponds to the GraspAndPlace task. This environment consists of dual-hands, an object and a bucket that requires us to pick up the object and put it into the bucket \\ \verb+1[|block - bucket| < 0.2]+ \\[1px]
    \hline
    BlockStack (428, 52) \\ This class corresponds to the Block Stack task. This environment involves dual hands and two blocks, and we need to stack the block as a tower \\ \verb|1[goal_dist_1 < 0.07 and goal_dist_2 < 0.07 and| \\ \verb|50 * (0.05 - z_dist_1) > 1]| \\[1px]
    \hline
    PushBlock (428, 52) \\ This class corresponds to the PushBlock task. This environment involves two hands and two blocks, we need to use both hands to reach and push the block to the desired goal separately. This is a  relatively simple task \\ \verb|1[left_dist <= 0.1 and right_dist <= 0.1] +| \\ \verb|0.5 * 1[left_dist <= 0.1 and right_dist > 0.1]| \\[1px]
    \hline
    TwoCatchUnderarm (446, 52) \\ This class corresponds to the TwoCatchUnderarm task. This environment is similar to Catch Underarm,  but with an object in each hand and the corresponding goal on the other hand. Therefore, the  environment requires two objects to be thrown into the other hand at the same time, which requires  a higher manipulation technique than the environment of a single object \\ \verb|1[goal_dist_1 + goal_dist_2 < 0.06]| \\[1px]
    \bottomrule
\end{tabularx}

\section{Baseline Details}
\label{appendix:baseline-details}
Language-to-Rewards (L2R) uses an LLM to generate a motion description from a natural language instruction and a set of reward API calls from the motion description. The reward is computed as the sum of outputs from the reward API calls. While the LLM automates the process of breaking down the task into basic low-level instructions, manual effort is still required to specify the motion description template, low-level reward API, and the API's function implementations.

All three parts require significant design considerations and can drastically affect L2R's performance and capabilities. Unfortunately, this makes comparison difficult since L2R requires manual engineering whereas Eureka is fully automatic—ambiguity thus arises from how much human-tuning should be done with L2R's components. Nonetheless, we seek to provide a fair comparison and base our implementation off two factors:
\begin{itemize}[leftmargin=*]
\item To create our motion description template, we reference L2R's quadruped and dexterous manipulator templates. Specifically, our templates consist of statements that set parameters to quantitative values and statements that relate two parameters. We also aim to mimic the style of L2R's template statements in general.
\item The reward API is designed so that each template statement can be faithfully written in terms of an API function. The functions are implemented to resemble their respective human reward terms from their environment; thus, L2R is given an advantage in that its components resemble the manually-tuned human reward. In a few exceptions where the human reward differs significantly from the L2R template style, we base our API implementation on the formulas provided in the L2R appendix.
\end{itemize}

L2R was designed to allow for an agent in a single environment to perform multiple tasks. Thus, each environment has its own motion description template and reward API. Since our experiments range over many agents and environments, we have one template and API for each Isaac task, and we generalize all Dexterity tasks into one environment with all necessary objects.

For illustration, our descriptor and coder prompts for the Dexterity experiments are below.

\renewcommand\lstlistingname{Prompt}
\setcounter{lstlisting}{0}

\begin{lstlisting}[basicstyle=\fontfamily{\ttdefault}\scriptsize, breaklines=true,caption={Dexterity descriptor prompt}]
We have two dexterous manipulators (shadow hands) and we want you to help plan how it should move to perform tasks using the following template:

[start of description]
object1={CHOICE: <INSERT OBJECTS HERE>} should be {CHOICE: close to, far from} object2={CHOICE: <INSERT OBJECTS HERE>, nothing}.
[optional] object3={CHOICE: <INSERT OBJECTS HERE>} should be {CHOICE: close to, far from} object4={CHOICE: <INSERT OBJECTS HERE>, nothing}.
[optional] object1 needs to have a rotation orientation similar to object2.
[optional] object3 needs to have a rotation orientation similar to object4.
<INSERT OPTIONAL HAND DESCRIPTIONS HERE>
[optional] doors needs to be {CHOICE: open, closed} {CHOICE: inward, outward}.
[optional] scissor needs to be opened to [NUM: 0.0] radians.
[optional] block2 needs to be stacked on top of block1.
[end of description]

Rules:
1. If you see phrases like [NUM: default_value], replace the entire phrase with a numerical value.
2. If you see phrases like {CHOICE: choice1, choice2, ...}, it means you should replace the entire
phrase with one of the choices listed.
3. If you see [optional], it means you only add that line if necessary for the task, otherwise remove that line.
4. The environment contains <INSERT OBJECTS HERE>. Do not invent new objects not listed here.
5. I will tell you a behavior/skill/task that I want the manipulator to perform and you will provide the full plan, even if you may only need to change a few lines. Always start the description with [start of plan] and end it with [end of plan].
6. You can assume that the hands are capable of doing anything, even for the most challenging task.
7. Your plan should be as close to the provided template as possible. Do not include additional details.
\end{lstlisting}

\begin{lstlisting}[basicstyle=\fontfamily{\ttdefault}\scriptsize, breaklines=true,caption={Dexterity coder prompt}]
We have a plan of a robot arm with palm to manipulate objects and we want you to turn that into the corresponding program with following functions:

```
def set_min_l2_distance_reward(name_obj_A, name_obj_B)
```
This term sets a reward for minimizing l2 distance between name_obj_A and name_obj_B so they get closer to each other.
name_obj_A and name_obj_B are selected from [<INSERT FIELDS HERE>].

```
def set_max_l2_distance_reward(name_obj_A, name_obj_B)
```
This term sets a reward for maximizing l2 distance between name_obj_A and name_obj_B so they get closer to each other.
name_obj_A and name_obj_B are selected from [<INSERT FIELDS HERE>].

```
def set_obj_orientation_reward(name_obj_A, name_obj_B)
```
This term encourages the orientation of name_obj_A to be close to the orientation of name_obj_B. name_obj_A and name_obj_B are selected from [<INSERT ORIENTATION FIELDS HERE>].

Example plan:
object1=object1 should be close to object2=object1_goal.
object1 needs to have a rotation orientation similar to object2.
To perform this task, the left manipulator's palm should move close to object1.

Example answer code:
```
set_min_l2_distance_reward("object1", "object1_goal")
set_min_l2_distance_reward("object1", "left_palm")
set_obj_orientation_reward("object1", "object1_goal")
```

Remember:
1. Always format the code in code blocks.
2. Do not wrap your code in a function. Your output should only consist of function calls like the example above.
3. Do not invent new functions or classes. The only allowed functions you can call are the ones listed above, and do not implement them. Do not leave unimplemented code blocks in your response.
4. The only allowed library is numpy. Do not import or use any other library.
5. If you are not sure what value to use, just use your best judge. Do not use None for anything.
6. Do not calculate the position or direction of any object (except for the ones provided above). Just use a number directly based on your best guess.
7. You do not need to make the robot do extra things not mentioned in the plan such as stopping the robot.
\end{lstlisting}

For the sections surrounded by angle brackets \verb|<>|, we specify a list of valid objects for each Dexterity task. For example, ShadowHandPen's list of objects is defined as follows:
\begin{lstlisting}[basicstyle=\fontfamily{\ttdefault}\scriptsize]
"shadow_hand_pen": ["left_palm", "right_palm", "left_forefinger", "left_middlefinger", "left_ringfinger", "left_littlefinger", "left_thumb", "right_forefinger", "right_middlefinger", "right_ringfinger", "right_littlefinger", "right_thumb", "pen_cap", "pen"]
\end{lstlisting}

A summary of terms and their implementations for each experiment is in Table ~\ref{table:l2r_form}. Note that many environments automatically randomize their target parameters during training after a reset or success criteria is met, which L2R cannot account for during the reward generation stage. Thus, while L2R's experiments define targets in terms of quantitative values, it's incompatible with our environments, and we define targets instead as relations between two parameters (usually the object and the object's target). 

\begin{table}[h] %
	\begin{center}
		\begin{tabular}{cc}
			\toprule
                \textbf{Reward Term} & \textbf{Formulation} \\
                \hline
                \textbf{Dexterity} & \\
                Minimize distance & $-\Vert p_1 - p_2 \Vert_2$ \\
                Maximize distance & $\Vert p_1 - p_2 \Vert_2$ \\
                Minimize orientation & $2\arcsin(\min(\Vert v(q_1\bar{q_2}) \Vert_2, 1))$ \\
                \hline
                \textbf{AllegroHand} & \\
                Minimize distance & $-\Vert p_1 - p_2 \Vert_2$ \\
                Maximize distance & $\Vert p_1 - p_2 \Vert_2$ \\
                Minimize orientation difference & $1/(\vert2\arcsin(\min(\Vert v(q_1\bar{q_2}) \Vert_2, 1))\vert+\epsilon)$ \\
                Maximize orientation difference & $-1/(\vert2\arcsin(\min(\Vert v(q_1\bar{q_2}) \Vert_2, 1))\vert+\epsilon)$ \\
                \hline
                \textbf{Ant} & \\
                Torso height & $-\vert h - h_t \vert$ \\
                Torso velocity & $-\vert \Vert v_{xy} \Vert_2 - v_t \vert$ \\
                Angle to target & $-\vert \theta - \theta_t \vert$ \\
                \hline
                \textbf{Anymal} & \\
                Minimize difference & $\exp -(x - x_t)^2$ \\
                \hline
                \textbf{BallBalance} & \\
                Ball position & $1 / (1 + \Vert p - p_t \Vert_2)$ \\
                Ball velocity & $1 / (1 + \Vert v - v_t \Vert_2)$ \\
                \hline
                \textbf{Cartpole} & \\
                Pole angle & $-(\theta - \theta_t)^2$ \\
                Pole velocity & $-\vert v - v_t \vert$ \\
                Cart velocity & $-\vert v - v_t \vert$ \\
                \hline
                \textbf{FrankaCabinet} & \\
                Minimize hand distance & $-\Vert p_1 - p_2 \Vert_2$ \\
                Maximize hand distance & $\Vert p_1 - p_2 \Vert_2$ \\
                Drawer extension & $-\vert p - p_t \vert$ \\
                \hline
                \textbf{Humanoid} & \\
                Torso height & $-\vert h - h_t \vert$ \\
                Torso velocity & $-\vert \Vert v_{xy} \Vert_2 - v_t \vert$ \\
                Angle to target & $-\vert \theta - \theta_t \vert$ \\
                \hline
                \textbf{Quadcopter} & \\
                Quadcopter position & $1 / (1 + \Vert p - p_t \Vert_2^2)$ \\
                Upright alignment & $1 / (1 + \vert 1 - n_z \vert^2)$ \\
                Positional velocity & $1 / (1 + \Vert v - v_t \Vert_2^2)$ \\
                Angular velocity & $1 / (1 + \Vert \omega - \omega_t \Vert_2^2)$ \\
                \hline
                \textbf{ShadowHand} & \\
                Minimize distance & $-\Vert p_1 - p_2 \Vert_2$ \\
                Maximize distance & $\Vert p_1 - p_2 \Vert_2$ \\
                Minimize orientation difference & $1/(\vert2\arcsin(\min(\Vert v(q_1\bar{q_2}) \Vert_2, 1))\vert+\epsilon)$ \\
                Maximize orientation difference & $-1/(\vert2\arcsin(\min(\Vert v(q_1\bar{q_2}) \Vert_2, 1))\vert+\epsilon)$ \\
			\bottomrule
		\end{tabular}
	\end{center}
	\caption{L2R reward primitives and their implementations. $v(q)$ denotes the vector part of quaternion $q$, subscript $t$ denotes target value, and $n$ denotes the normal vector (orientation). All components are weighed equally.}
    \label{table:l2r_form}
\end{table}

\subsection{L2R Reward Examples}

\renewcommand\lstlistingname{Example}
\setcounter{lstlisting}{0}

\begin{lstlisting}[language=python, basicstyle=\fontfamily{\ttdefault}\scriptsize, caption={L2R reward function on Humanoid, Human Normalized Score: 0.0}]
set_torso_height_reward(1.1)
set_torso_velocity_reward(3.6)
set_angle_to_target_reward(0.0)
\end{lstlisting}
\begin{lstlisting}[language=python, basicstyle=\fontfamily{\ttdefault}\scriptsize, caption={L2R reward function on ShadowHandKettle, Success Rate: 0.07}]
set_min_l2_distance_reward("kettle_handle", "bucket_handle")
set_min_l2_distance_reward("kettle_spout", "bucket_handle")
set_min_l2_distance_reward("left_palm", "bucket_handle")
set_min_l2_distance_reward("right_palm", "kettle_handle")
set_min_l2_distance_reward("left_thumb", "bucket_handle")
set_min_l2_distance_reward("right_thumb", "kettle_handle")
\end{lstlisting}

\section{\ourmethod Details}
\label{appendix:experimental-details}

\textbf{Environment as Context.} In Isaac Gym, the simulator adopts a environment design pattern in which the environment observation code is typically written inside a \texttt{compute\_observations()} function within the environment object class; this applies to all our environments. Therefore, we have written an automatic script to extract just the observation portion of the environment source code. This is done largely to reduce our experiment cost as longer context induces higher cost. Furthermore, given that current LLMs have context length limit, this task agnostic way of trimming the environment code before feeing it to the context allows us to fit every environment source code into context. 

\renewcommand\lstlistingname{Example}
\setcounter{lstlisting}{0}
\begin{lstlisting}[language=python, basicstyle=\fontfamily{\ttdefault}\scriptsize, caption={Humanoid environment observation given to \ourmethod.}]
class Humanoid(VecTask):
    """Rest of the environment definition omitted."""
    def compute_observations(self):
        self.gym.refresh_dof_state_tensor(self.sim)
        self.gym.refresh_actor_root_state_tensor(self.sim)

        self.gym.refresh_force_sensor_tensor(self.sim)
        self.gym.refresh_dof_force_tensor(self.sim)
        self.obs_buf[:], self.potentials[:], self.prev_potentials[:], self.up_vec[:], self.heading_vec[:] = compute_humanoid_observations(
            self.obs_buf, self.root_states, self.targets, self.potentials,
            self.inv_start_rot, self.dof_pos, self.dof_vel, self.dof_force_tensor,
            self.dof_limits_lower, self.dof_limits_upper, self.dof_vel_scale,
            self.vec_sensor_tensor, self.actions, self.dt, self.contact_force_scale, self.angular_velocity_scale,
            self.basis_vec0, self.basis_vec1)

def compute_humanoid_observations(obs_buf, root_states, targets, potentials, inv_start_rot, dof_pos, dof_vel,
                                  dof_force, dof_limits_lower, dof_limits_upper, dof_vel_scale,
                                  sensor_force_torques, actions, dt, contact_force_scale, angular_velocity_scale,
                                  basis_vec0, basis_vec1):
    # type: (Tensor, Tensor, Tensor, Tensor, Tensor, Tensor, Tensor, Tensor, Tensor, Tensor, float, Tensor, Tensor, float, float, float, Tensor, Tensor) -> Tuple[Tensor, Tensor, Tensor, Tensor, Tensor]

    torso_position = root_states[:, 0:3]
    torso_rotation = root_states[:, 3:7]
    velocity = root_states[:, 7:10]
    ang_velocity = root_states[:, 10:13]

    to_target = targets - torso_position
    to_target[:, 2] = 0

    prev_potentials_new = potentials.clone()
    potentials = -torch.norm(to_target, p=2, dim=-1) / dt

    torso_quat, up_proj, heading_proj, up_vec, heading_vec = compute_heading_and_up(
        torso_rotation, inv_start_rot, to_target, basis_vec0, basis_vec1, 2)

    vel_loc, angvel_loc, roll, pitch, yaw, angle_to_target = compute_rot(
        torso_quat, velocity, ang_velocity, targets, torso_position)

    roll = normalize_angle(roll).unsqueeze(-1)
    yaw = normalize_angle(yaw).unsqueeze(-1)
    angle_to_target = normalize_angle(angle_to_target).unsqueeze(-1)
    dof_pos_scaled = unscale(dof_pos, dof_limits_lower, dof_limits_upper)

    obs = torch.cat((torso_position[:, 2].view(-1, 1), vel_loc, angvel_loc * angular_velocity_scale,
                     yaw, roll, angle_to_target, up_proj.unsqueeze(-1), heading_proj.unsqueeze(-1),
                     dof_pos_scaled, dof_vel * dof_vel_scale, dof_force * contact_force_scale,
                     sensor_force_torques.view(-1, 12) * contact_force_scale, actions), dim=-1)

    return obs, potentials, prev_potentials_new, up_vec, heading_vec



\end{lstlisting}

\textbf{\ourmethod Reward History.} Given that LLMs have limited context, we also trim the \ourmethod dialogue such that only the last reward and its reward reflection (in addition to initial system prompt) is kept in the context for the generation of the next reward. In other word, the reward improvement is \textit{Markovian}. This is standard in gradient-free optimization, and we find this simplification to work well in practice.

\textbf{\ourmethod Reward Evaluation.} All intermediate \ourmethod reward functions are evaluated using 1 PPO run with the default task parameters. The final \ourmethod reward, like all other baseline reward functions, are evaluated using 5 PPO runs with the average performance on the task fitness function $F$ as the reward performance. 

\textbf{Human Normalized Score Reporting.} Given that there are several significant outliers in human normalized score when \ourmethod is substantially better both Human and Sparse on a task, when reporting the average normalized improvement in our abstract, we adjust the score so that the normalized score must lie between $[0,3]$ per task before computing the average over all 29 tasks. 

\subsection{\arxiv{Pen Spinning Tasks}}
\label{appendix:eureka-pen-spinning}
\arxiv{The pen spinning environment builds on the original Shadow Hand environment in Isaac Gym without any changes to the physics parameters; the cube object in the original task is swapped out with the pen object, which is an unused asset in the Isaac Gym repository. In both the pre-training and fine-tuning stages, the task is to reorient a pen to a target 3D pose, which is provided as a goal to the policy. When the current target pose is deemed achieved (i.e., pose difference lower than certain threshold), then a new target pose is provided. In the pre-training stage, a random pose from SO(3) 3D orientation group is sampled uniformly; this mechanism is identical to the original benchmark cube reorientation task. During the fine-tuning stage, the target poses are a predetermined sequence of waypoints that specify the pen spinning patterns to be achieved. When the policy reaches the current waypoint, the target pose will switch to the next waypoint in the sequence. If the policy can consecutively reach all the waypoints, then it has accomplished one cycle of the pen spinning pattern. The waypoints will relay continuously, allowing the policy to spin the pen for as many cycles as the policy is capable of until the episode length is reached.}

\subsection{\ourmethod from Human Initialization}
\label{appendix:eureka-human-init-details}
In our human initialization experiments, we use Eureka to improve human-written reward functions. This can be done by modifying the first \ourmethod iteration to use the human reward in place of the LLM-generated one, thereby "assuming" that Eureka's first proposed reward is the human reward. To complete this iteration, we use the human reward in IsaacGym, compute feedback, and query the LLM to generate new reward functions based on the human reward and the reward reflections. Future iterations are identical to the default \ourmethod setting.

To provide the human reward in the first iteration, we refactor the code slightly to be consistent with the \ourmethod reward format, which exposes the individual reward components in a dictionary. Furthermore, as human reward functions are often written in less interpretable fashion than \ourmethod rewards (see App.~\ref{appendix:human-comparison-examples} for an example) , we also strip away excess variables and parameters besides those needed for the actual reward computation.

\subsection{\ourmethod from Human Feedback}
\label{appendix:eureka-rlhf-details}

In our human reward reflection experiment, we investigate whether humans can provide reward reflection for desideratum such as ``running with natural gait'' that may be difficult to express via a task fitness function. We repeat the \ourmethod procedure with the following modifications: (1) We only sample 1 reward per iteration, and (2) a human textual input will replace the automatically constructed reward reflection as in the main experiment. To ensure that the human textual input do not require domain expertise, we have used feedback that is as colloquial as possible; the full conversation is shown in App.~\ref{appendix:human-reward-reflection-example}.

After the \ourmethod-HF agent is trained, we have asked 20 unfamiliar users to indicate their preferences between two videos shown in random order, one depicting the \ourmethod-HF Humanoid agent and the other one depicting the original best \ourmethod agent. These 20 users are other graduate and undergraduate students that have a wide range of familiarity in reinforcement learning, but are not involved with this research.

\subsection{\arxiv{Computation Resources}}
\label{appendix:computation-resources}
\arxiv{All our experiments took place on a single 8 A100 GPU station. Each individual Eureka run took less than one day of wall clock time on our GPU station. Given that Eureka evaluates at most 16 reward functions at a time, and each RL run on Isaac Gym takes at most 8GB of GPU memory, Eureka can be run on 4 V100 GPUs, which is readily accessible on an academic compute budget.}

\section{\arxiv{\ourmethod on Mujoco Environments}}
\label{appendix:eureka-humanoid}
\arxiv{To fully test the generality of \ourmethod with regard to code syntax and physics simulation, we also evaluate \ourmethod on the OpenAI Gym Humanoid environment~\citep{brockman2016openai} implemented using Mujoco~\citep{todorov2012mujoco}. We make no changes to the prompts in App.~\ref{appendix:full-prompts} except the formatting tip that instructs the LLM to use numpy array instead of pytorch tensor when performing matrix operations in the generated program. The observation code for \ourmethod context is displayed in Example 1 below. As shown, compared to the observation code for the Isaac Gym Humanoid task (Example 1 in App.~\ref{appendix:experimental-details}), this variant conveys the observation information in a very different manner: a commented block reveals the state and action spaces, and the variables themselves have to be indexed from a monolithic observation array.}

The comparison against the official human-written reward function is in Tab.~\ref{table:eureka-mujoco-humanoid}. Despite vastly different observation space and code syntax, \ourmethod remains effective and generates reward functions that outperform the official human reward function. 
\begin{table}
\caption{\small{\ourmethod on Mujoco Humanoid environment. As in Isaac Gym Humanoid, the task fitness function is the Humanoid's forward velocity.}}
\label{table:eureka-mujoco-humanoid}
\centering
\resizebox{0.5\textwidth}{!}{
\begin{tabular}[c]{l|cc}\toprule
Task & \ourmethod & Human  \\ \midrule
Mujoco Humanoid& \textbf{7.68} & 5.92 \\
\bottomrule
\end{tabular}
}
\end{table}

\renewcommand\lstlistingname{Example}
\setcounter{lstlisting}{0}
\begin{lstlisting}[language=python, basicstyle=\fontfamily{\ttdefault}\scriptsize, caption={Mujoco Humanoid environment observation code given to \ourmethod.}]
class HumanoidEnv(MujocoEnv, utils.EzPickle):
    """Rest of the environment definition omitted."""

    """
    ### Observation Space

    Observations consist of positional values of different body parts of the Humanoid,
     followed by the velocities of those individual parts (their derivatives) with all the
     positions ordered before all the velocities.

    Byy default, the observation is a `ndarray` with shape `(376,)` where the elements correspond to the following:

    | Num | Observation                                                                                                     | Min  | Max | Name (in corresponding XML file) | Joint | Unit                       |
    | --- | -------------------------------------------------------------| ---- | --- | -------------------------------- | ----- | -------------------------- |
    | 0   | z-coordinate of the torso (centre)                                                                              | -Inf | Inf | root                             | free  | position (m)               |
    | 1   | x-orientation of the torso (centre)                                                                             | -Inf | Inf | root                             | free  | angle (rad)                |
    | 2   | y-orientation of the torso (centre)                                                                             | -Inf | Inf | root                             | free  | angle (rad)                |
    | 3   | z-orientation of the torso (centre)                                                                             | -Inf | Inf | root                             | free  | angle (rad)                |
    | 4   | w-orientation of the torso (centre)                                                                             | -Inf | Inf | root                             | free  | angle (rad)                |
    | 5   | z-angle of the abdomen (in lower_waist)                                                                         | -Inf | Inf | abdomen_z                        | hinge | angle (rad)                |
    | 6   | y-angle of the abdomen (in lower_waist)                                                                         | -Inf | Inf | abdomen_y                        | hinge | angle (rad)                |
    | 7   | x-angle of the abdomen (in pelvis)                                                                              | -Inf | Inf | abdomen_x                        | hinge | angle (rad)                |
    | 8   | x-coordinate of angle between pelvis and right hip (in right_thigh)                                             | -Inf | Inf | right_hip_x                      | hinge | angle (rad)                |
    | 9   | z-coordinate of angle between pelvis and right hip (in right_thigh)                                             | -Inf | Inf | right_hip_z                      | hinge | angle (rad)                |
    | 19  | y-coordinate of angle between pelvis and right hip (in right_thigh)                                             | -Inf | Inf | right_hip_y                      | hinge | angle (rad)                |
    | 11  | angle between right hip and the right shin (in right_knee)                                                      | -Inf | Inf | right_knee                       | hinge | angle (rad)                |
    | 12  | x-coordinate of angle between pelvis and left hip (in left_thigh)                                               | -Inf | Inf | left_hip_x                       | hinge | angle (rad)                |
    | 13  | z-coordinate of angle between pelvis and left hip (in left_thigh)                                               | -Inf | Inf | left_hip_z                       | hinge | angle (rad)                |
    | 14  | y-coordinate of angle between pelvis and left hip (in left_thigh)                                               | -Inf | Inf | left_hip_y                       | hinge | angle (rad)                |
    | 15  | angle between left hip and the left shin (in left_knee)                                                         | -Inf | Inf | left_knee                        | hinge | angle (rad)                |
    | 16  | coordinate-1 (multi-axis) angle between torso and right arm (in right_upper_arm)                                | -Inf | Inf | right_shoulder1                  | hinge | angle (rad)                |
    | 17  | coordinate-2 (multi-axis) angle between torso and right arm (in right_upper_arm)                                | -Inf | Inf | right_shoulder2                  | hinge | angle (rad)                |
    | 18  | angle between right upper arm and right_lower_arm                                                               | -Inf | Inf | right_elbow                      | hinge | angle (rad)                |
    | 19  | coordinate-1 (multi-axis) angle between torso and left arm (in left_upper_arm)                                  | -Inf | Inf | left_shoulder1                   | hinge | angle (rad)                |
    | 20  | coordinate-2 (multi-axis) angle between torso and left arm (in left_upper_arm)                                  | -Inf | Inf | left_shoulder2                   | hinge | angle (rad)                |
    | 21  | angle between left upper arm and left_lower_arm                                                                 | -Inf | Inf | left_elbow                       | hinge | angle (rad)                |
    | 22  | x-coordinate velocity of the torso (centre)                                                                     | -Inf | Inf | root                             | free  | velocity (m/s)             |
    | 23  | y-coordinate velocity of the torso (centre)                                                                     | -Inf | Inf | root                             | free  | velocity (m/s)             |
    | 24  | z-coordinate velocity of the torso (centre)                                                                     | -Inf | Inf | root                             | free  | velocity (m/s)             |
    | 25  | x-coordinate angular velocity of the torso (centre)                                                             | -Inf | Inf | root                             | free  | anglular velocity (rad/s)  |
    | 26  | y-coordinate angular velocity of the torso (centre)                                                             | -Inf | Inf | root                             | free  | anglular velocity (rad/s)  |
    | 27  | z-coordinate angular velocity of the torso (centre)                                                             | -Inf | Inf | root                             | free  | anglular velocity (rad/s)  |
    | 28  | z-coordinate of angular velocity of the abdomen (in lower_waist)                                                | -Inf | Inf | abdomen_z                        | hinge | anglular velocity (rad/s)  |
    | 29  | y-coordinate of angular velocity of the abdomen (in lower_waist)                                                | -Inf | Inf | abdomen_y                        | hinge | anglular velocity (rad/s)  |
    | 30  | x-coordinate of angular velocity of the abdomen (in pelvis)                                                     | -Inf | Inf | abdomen_x                        | hinge | aanglular velocity (rad/s) |
    | 31  | x-coordinate of the angular velocity of the angle between pelvis and right hip (in right_thigh)                 | -Inf | Inf | right_hip_x                      | hinge | anglular velocity (rad/s)  |
    | 32  | z-coordinate of the angular velocity of the angle between pelvis and right hip (in right_thigh)                 | -Inf | Inf | right_hip_z                      | hinge | anglular velocity (rad/s)  |
    | 33  | y-coordinate of the angular velocity of the angle between pelvis and right hip (in right_thigh)                 | -Inf | Inf | right_hip_y                      | hinge | anglular velocity (rad/s)  |
    | 34  | angular velocity of the angle between right hip and the right shin (in right_knee)                              | -Inf | Inf | right_knee                       | hinge | anglular velocity (rad/s)  |
    | 35  | x-coordinate of the angular velocity of the angle between pelvis and left hip (in left_thigh)                   | -Inf | Inf | left_hip_x                       | hinge | anglular velocity (rad/s)  |
    | 36  | z-coordinate of the angular velocity of the angle between pelvis and left hip (in left_thigh)                   | -Inf | Inf | left_hip_z                       | hinge | anglular velocity (rad/s)  |
    | 37  | y-coordinate of the angular velocity of the angle between pelvis and left hip (in left_thigh)                   | -Inf | Inf | left_hip_y                       | hinge | anglular velocity (rad/s)  |
    | 38  | angular velocity of the angle between left hip and the left shin (in left_knee)                                 | -Inf | Inf | left_knee                        | hinge | anglular velocity (rad/s)  |
    | 39  | coordinate-1 (multi-axis) of the angular velocity of the angle between torso and right arm (in right_upper_arm) | -Inf | Inf | right_shoulder1                  | hinge | anglular velocity (rad/s)  |
    | 40  | coordinate-2 (multi-axis) of the angular velocity of the angle between torso and right arm (in right_upper_arm) | -Inf | Inf | right_shoulder2                  | hinge | anglular velocity (rad/s)  |
    | 41  | angular velocity of the angle between right upper arm and right_lower_arm                                       | -Inf | Inf | right_elbow                      | hinge | anglular velocity (rad/s)  |
    | 42  | coordinate-1 (multi-axis) of the angular velocity of the angle between torso and left arm (in left_upper_arm)   | -Inf | Inf | left_shoulder1                   | hinge | anglular velocity (rad/s)  |
    | 43  | coordinate-2 (multi-axis) of the angular velocity of the angle between torso and left arm (in left_upper_arm)   | -Inf | Inf | left_shoulder2                   | hinge | anglular velocity (rad/s)  |
    | 44  | angular velocitty of the angle between left upper arm and left_lower_arm                                        | -Inf | Inf | left_elbow                       | hinge | anglular velocity (rad/s)  |

    Additionally, after all the positional and velocity based values in the table,
    the observation contains (in order):
    - *cinert:* Mass and inertia of a single rigid body relative to the center of mass
    (this is an intermediate result of transition). It has shape 14*10 (*nbody * 10*)
    and hence adds to another 140 elements in the state space.
    - *cvel:* Center of mass based velocity. It has shape 14 * 6 (*nbody * 6*) and hence
    adds another 84 elements in the state space
    - *qfrc_actuator:* Constraint force generated as the actuator force. This has shape
    `(23,)`  *(nv * 1)* and hence adds another 23 elements to the state space.
    - *cfrc_ext:* This is the center of mass based external force on the body.  It has shape
    14 * 6 (*nbody * 6*) and hence adds to another 84 elements in the state space.
    where *nbody* stands for the number of bodies in the robot and *nv* stands for the
    number of degrees of freedom (*= dim(qvel)*)

    The (x,y,z) coordinates are translational DOFs while the orientations are rotational
    DOFs expressed as quaternions.

    Note that the obs argument passed to the reward function is a concatenation of obs across all environments, so it will have shape (n, 376).
    """

    def _get_obs(self):
        position = self.data.qpos.flat.copy()
        velocity = self.data.qvel.flat.copy()

        com_inertia = self.data.cinert.flat.copy()
        com_velocity = self.data.cvel.flat.copy()

        actuator_forces = self.data.qfrc_actuator.flat.copy()
        external_contact_forces = self.data.cfrc_ext.flat.copy()

        if self._exclude_current_positions_from_observation:
            position = position[2:]

        return np.concatenate(
            (
                position,
                velocity,
                com_inertia,
                com_velocity,
                actuator_forces,
                external_contact_forces,
            )
        )
\end{lstlisting}

\section{Additional Results}
\label{appendix:additional-results}

\arxiv{\textbf{Reinforcement Learning Training Curves.} In Fig.~\ref{fig:eureka-dexterity-training_curves}, we present the aggregate RL training curves on the Dexterity benchmark. As shown, \ourmethod reward functions enjoy improved sample efficiency compared to all baseline rewards. We do notice that \ourmethod reward functions, on average, exhibits less smooth trend than the Human reward functions. We hypothesize that this is due to the fact that the underlying RL algorithm is not tuned for the \ourmethod rewards; instead, they are tuned for the Human reward functions.}

\begin{minipage}{\linewidth}
  \centering
  \includegraphics[width=0.5\linewidth]{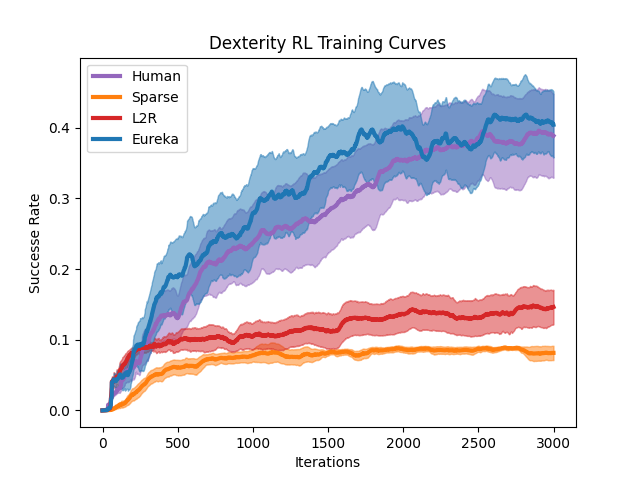}
  \captionof{figure}{\arxiv{\ourmethod reward functions enjoy improved sample efficiency compared to various baseline reward functions on aggregate over 20 Dexterity tasks.}}
  \label{fig:eureka-dexterity-training_curves}
\end{minipage}

\arxiv{\textbf{Additional Evaluation Metrics.} In Fig.~\ref{fig:eureka-rliable}, we present holistic evaluation metrics, such as mean, median, and interquantile mean, as suggested by~\citet{agarwal2021deep}. All metrics and associated 95\% confidence intervals are computed over the set of 100 RL runs from 5 RL runs for each method's final reward function on all 20 Dexterity tasks. As shown, on all evaluation metrics, \ourmethod is consistently effective and outperforms all baselines.}

\arxiv{Furthermore, using the same set of runs, we also compute the probability that \ourmethod outperforms the baselines with 95\% confidence intervals. The results are shown in Fig.~\ref{fig:eureka-probabilities}. As shown, our experimental results provide strong evidence that \ourmethod's improvement over baselines are statistically significant.}

\begin{minipage}{\linewidth}
  \centering
  \includegraphics[width=\linewidth]{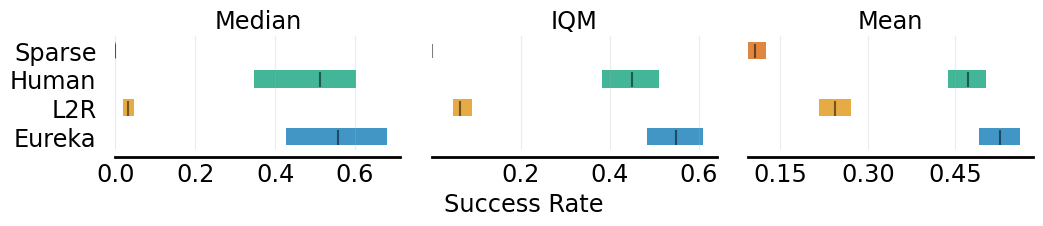}
  \captionof{figure}{\ourmethod reward functions are more effective in aggregate under various evaluation metrics.}
    \label{fig:eureka-rliable}

    \includegraphics[width=0.45\linewidth]{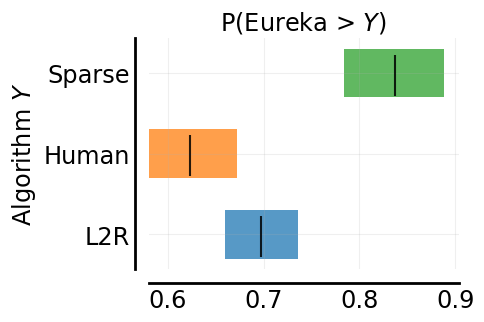}
  \captionof{figure}{\ourmethod reward functions' improvement over alternative reward functions are statistically significant.}
  \label{fig:eureka-probabilities}
\end{minipage}

\textbf{Dexterity Performance Breakdown.} We present the raw success rates of \ourmethod, L2R, Human, and Sparse in Fig.~\ref{fig:eureka-dexterity-per-task}.

\begin{minipage}{\linewidth}
  \centering
  \includegraphics[width=\linewidth]{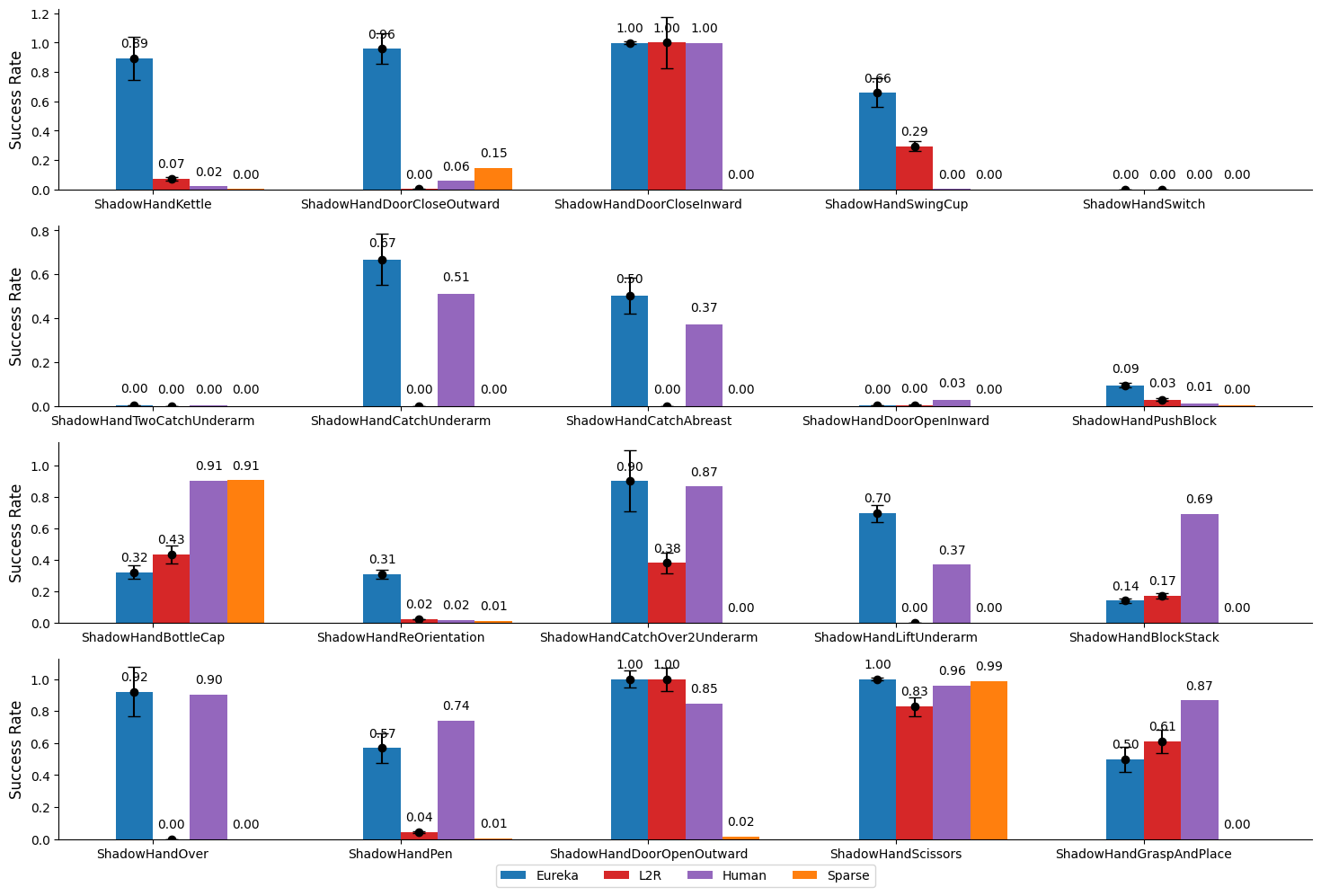}
  \captionof{figure}{Raw success rates of all methods on the Dexterity benchmark.}
  \label{fig:eureka-dexterity-per-task}
\end{minipage}

\textbf{Reward Reflection Ablations.} In Fig.~\ref{fig:eureka-verbalization-ablation}, we provide a detailed per-task breakdown on the impact of removing reward reflection in the \ourmethod feedback. In this ablation, we are interested in the \textit{average} human normalized score over independent \ourmethod restarts because the average is more informative than the max (the metric used in all other experiments) in revealing LLM behavior change on aggregate. As shown, removing reward reflection generally has a negative impact on the reward performance. The deterioration is more pronounced for high-dimensional tasks, demonstrating that reward reflection indeed can provide targeted reward editing that is more instrumental for difficult tasks that require many state components to interact in the reward functions. 

\begin{minipage}{\linewidth}
  \centering
  \includegraphics[width=\linewidth]{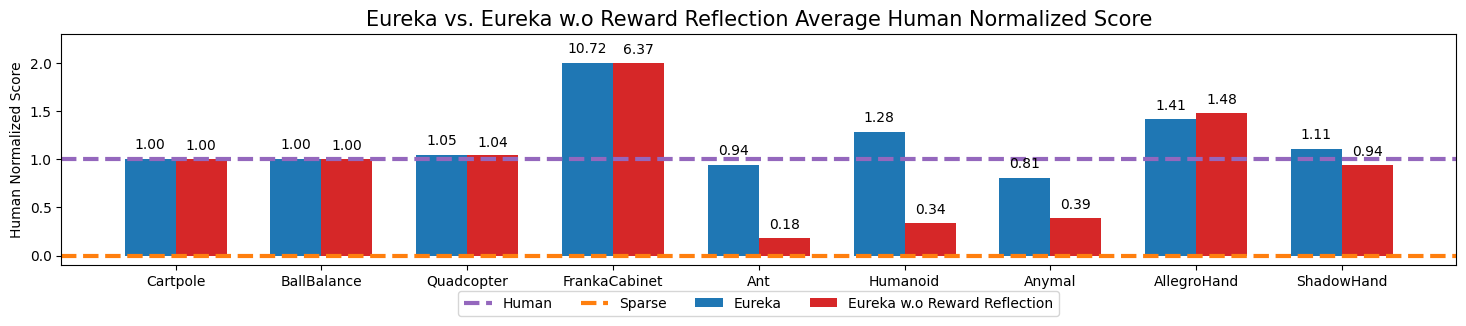}
  \captionof{figure}{\ourmethod without the reward reflection mechanism exhibits degraded performance.}
  \label{fig:eureka-verbalization-ablation}
\end{minipage}

\textbf{\ourmethod with GPT-3.5.} In Fig.~\ref{fig:eureka-gpt35-ablation}, we compare the performance of \ourmethod with GPT-4 (the original one reported in the paper) and \ourmethod with GPT-3.5; specifically, we use \texttt{gpt-3.5-turbo-16k-0613} in the OpenAI API. While the absolute performance goes down, we see that \ourmethod (GPT-3.5) still performs comparably and exceeds human-engineered rewards on the dexterous manipulation tasks. These results suggest that the \ourmethod principles are general and can be also applied to less performant base coding LLMs.

\begin{minipage}{\linewidth}
  \centering
  \includegraphics[width=\linewidth]{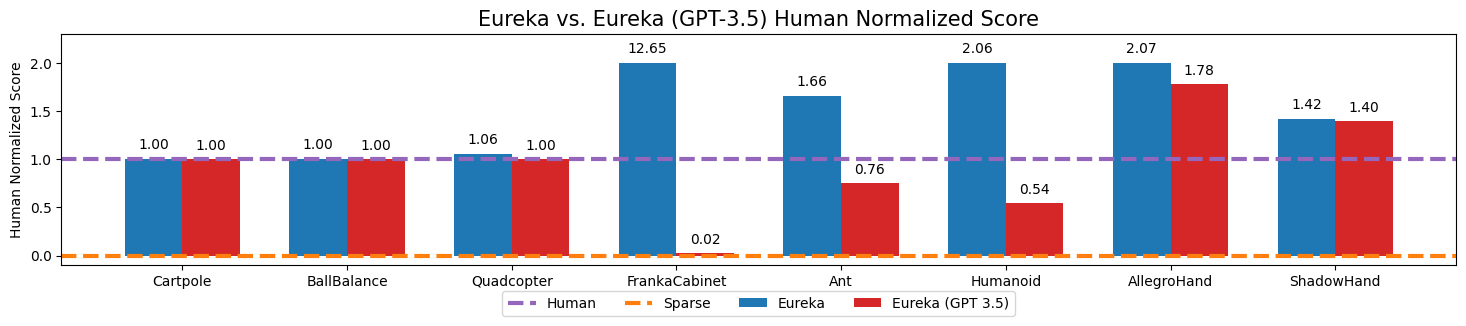}
  \captionof{figure}{Using GPT3.5 observes performance degradation in \ourmethod but still remains comparable to GPT-4 on a majority of the tasks.}
  \label{fig:eureka-gpt35-ablation}
\end{minipage}

\textbf{Reward Correlation Experiments.} To provide a more bird-eye view comparison against human rewards, we assess the novelty of \ourmethod rewards. Given that programs that syntactically differ may functionally be identical, we propose to evaluate the Pearson \textit{correlation} between \ourmethod and human rewards on all the Isaac task. These tasks are ideal for this test because many of them have been widely used in RL research, even if the IsaacGym implementation may not have been seen in GPT-4 training, so it is possible that \ourmethod produces rewards that are merely cosmetically different.  To do this, for a given policy training run using a \ourmethod reward, we gather all training transitions and compute their respectively \ourmethod and human reward values, which can then be used to compute their correlation. Then, we plot the correlation against the human normalized score on a scatter-plot. The resulting scatter-plot is displayed in Fig.~\ref{fig:eureka-correlation}.
In Fig.~\ref{fig:eureka-correlation-by-task}, we also provide the average correlation per task. As shown, as the tasks become more high-dimensional and harder to solve, the correlations exhibit a downward trend. This validates our hypothesis that the harder the task is, the less optimal the human rewards are, and consequently more room for \ourmethod to generate truly novel and different rewards. 

\begin{minipage}{\linewidth}
  \centering
  \includegraphics[width=0.8\linewidth]{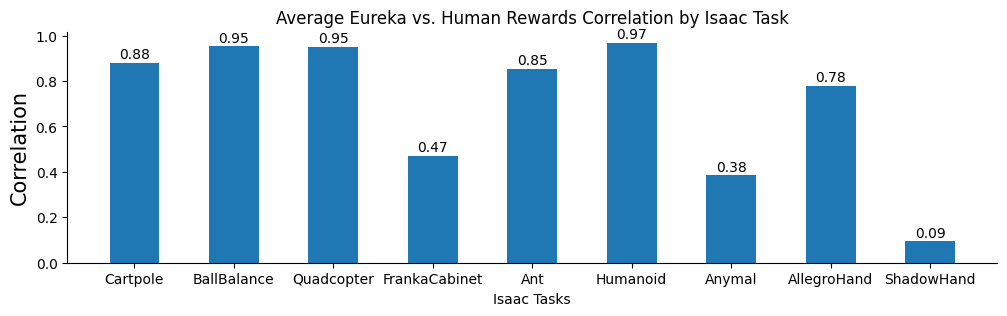}
  \captionof{figure}{\ourmethod rewards are less correlated with human rewards when the tasks are more high-dimensional and less common in the reinforcement learning literature.}
  \label{fig:eureka-correlation-by-task}
\end{minipage}

\section{\ourmethod Reward Examples}
\label{appendix:examples}
In this section, we provide several unmodified \ourmethod reward examples from various experiments we detailed in the main paper. 

\subsection{Reward Reflection Examples.}
\label{appendix:reward-reflection-examples}
We illustrate \ourmethod reward reflection step by displaying the sequence of previous \ourmethod reward, \ourmethod reward reflection, and then the subsequent \ourmethod reward. 

\renewcommand\lstlistingname{Example}
\setcounter{lstlisting}{0}

\begin{lstlisting}[language=python, basicstyle=\fontfamily{\ttdefault}\scriptsize, caption={\ourmethod Reward Reflection on ShadowHand. Iteration 1 Score: 9.29, Iteration 2 Score: 10.43}]
Iteration 1: Eureka Output:
```python
import torch
from typing import Tuple, Dict

@torch.jit.script
def compute_reward(object_rot: torch.Tensor, goal_rot: torch.Tensor) -> Tuple[torch.Tensor, Dict[str, torch.Tensor]]:
    rot_diff = torch.abs(torch.sum(object_rot * goal_rot, dim=1) - 1) / 2
    rotation_reward = torch.exp(-20 * rot_diff)
    
    # Scaling factor for the rotation reward
    rotation_temp = 20.0
    
    total_reward = rotation_reward
    
    reward_components = {
        "rotation_reward": rotation_reward
    }
    
    return total_reward, reward_components
```
This reward function computes the difference between the current object rotation and the target rotation. It calculates a rotation_reward based on the absolute difference, scaled by a temperature parameter (rotation_temp). The total reward is the rotation_reward.

#########################################################################

Iteration 1: Reward Reflection:
We trained a RL policy using the provided reward function code and tracked the values of the individual components in the reward function as well as global policy metrics such as success rates and episode lengths after every 300 epochs and the maximum, mean, minimum values encountered:
rotation_reward: ['0.03', '0.31', '0.30', '0.32', '0.34', '0.34', '0.34', '0.34', '0.34', '0.32'], Max: 0.36, Mean: 0.32, Min: 0.03 
success_rate: ['0.00', '0.83', '1.85', '2.89', '3.72', '4.94', '5.99', '6.90', '7.71', '8.83'], Max: 9.29, Mean: 4.81, Min: 0.00 
episode_lengths: ['7.07', '384.30', '378.22', '378.06', '406.11', '415.43', '402.64', '412.50', '384.10', '434.24'], Max: 482.35, Mean: 396.02, Min: 7.07

Please carefully analyze the policy feedback and provide a new, improved reward function that can better solve the task. Some helpful tips for analyzing the policy feedback:
    (1) If the success rates are always near zero, then you must rewrite the entire reward function
    (2) If the values for a certain reward component are near identical throughout, then this means RL is not able to optimize this component as it is written. You may consider
        (a) Changing its scale or the value of its temperature parameter
        (b) Re-writing the reward component 
        (c) Discarding the reward component
    (3) If some reward component magnitude is significantly larger, then you must re-scale its value to a proper range
    
Please analyze each existing reward component in the suggested manner above first, and then write the reward function code. The output of the reward function should consist of two items:
    (1) the total reward,
    (2) a dictionary of each individual reward component.
The code output should be formatted as a python code string: "```python ... ```".

Some helpful tips for writing the reward function code:
    (1) You may find it helpful to normalize the reward to a fixed range by applying transformations like torch.exp to the overall reward or its components
    (2) If you choose to transform a reward component, then you must also introduce a temperature parameter inside the transformation function; this parameter must be a named variable in the reward function and it must not be an input variable. Each transformed reward component should have its own temperature variable
    (3) Make sure the type of each input variable is correctly specified; a float input variable should not be specified as torch.Tensor
    (4) Most importantly, the reward code input variables must contain only attributes of the provided environment class definition (namely, variables that have prefix self.). Under no circumstance can you introduce new input variables.

...

#########################################################################

Iteration 2: Eureka Output:
Based on the analysis of the policy feedback, the rotation_reward component has minimum, maximum, and mean values within a fairly close range and seems stable. The success_rate has increased gradually, indicating that the agent is learning to some degree. However, the episode_lengths have quite some variance, which could indicate that the agent does not consistently perform well during training.

Since the rotation_reward component seems mostly adequate, I would recommend introducing another component, such as a penalty for large angular velocities, to reinforce stable spinning behavior. Here is a new reward function that incorporates an angular_velocity_penalty:

```python
import torch
from typing import Tuple, Dict

@torch.jit.script
def compute_reward(object_rot: torch.Tensor, goal_rot: torch.Tensor, object_angvel: torch.Tensor) -> Tuple[torch.Tensor, Dict[str, torch.Tensor]]:
    # Rotation reward
    rot_diff = torch.abs(torch.sum(object_rot * goal_rot, dim=1) - 1) / 2
    rotation_reward = torch.exp(-20 * rot_diff)
    rotation_temp = 20.0

    # Angular velocity penalty
    angvel_norm = torch.norm(object_angvel, dim=1)
    angvel_threshold = 5.0
    angvel_penalty_temp = 10.0
    angular_velocity_penalty = torch.where(angvel_norm > angvel_threshold, torch.exp(-angvel_penalty_temp * (angvel_norm - angvel_threshold)), torch.zeros_like(angvel_norm))

    total_reward = rotation_reward - angular_velocity_penalty
    
    reward_components = {
        "rotation_reward": rotation_reward,
        "angular_velocity_penalty": angular_velocity_penalty
    }
    
    return total_reward, reward_components
```

The new reward function applies a penalty for large angular velocities (larger than the angvel_threshold) using a temperature parameter (angvel_penalty_temp) to scale the penalty. The total reward now combines the rotation_reward with the angular_velocity_penalty.
\end{lstlisting}
\begin{lstlisting}[language=python, basicstyle=\fontfamily{\ttdefault}\scriptsize, caption={\ourmethod Reward Reflection on Humanoid. Iteration 2 Score: 3.79, Iteration 3 Score: 4.37}]
Iteration 2: Eureka Output:
Based on the provided feedback, it seems like the `track_reward` component is not contributing to the learning process as all its values are 0. Therefore, I will discard the `track_reward` and write a new reward function that emphasizes the `velocity_reward` component. I will also add a new component to the reward function to encourage the agent to maintain upright posture.

Below is the updated reward function:

```python
import torch
from typing import Tuple, Dict

@torch.jit.script
def compute_reward(root_states: torch.Tensor, dt: float) -> Tuple[torch.Tensor, Dict[str, torch.Tensor]]:
    # Calculate the forward velocity
    forward_velocity = root_states[:, 7]
    
    # Reward for forward velocity
    velocity_temperature = 1.0  # You can tune this temperature parameter to adjust how much weight the forward velocity gets
    velocity_reward = torch.exp(forward_velocity / velocity_temperature)
    
    # Calculate the height of the humanoid torso
    torso_height = root_states[:, 2]

    # Reward for maintaining upright posture
    posture_temperature = 1.0  # Tune this temperature parameter to adjust the weight of the posture reward component
    posture_reward = torch.exp(torso_height / posture_temperature)
    
    # Combine rewards
    total_reward = velocity_reward + posture_reward
    
    reward_components = {'velocity_reward': velocity_reward, 'posture_reward': posture_reward}
    
    return total_reward, reward_components
```

I have removed the `track_reward` component and added a new component, `posture_reward`, to encourage the humanoid to maintain an upright posture. The posture reward is computed as the exponential of the humanoid's torso height divided by a temperature parameter. Adjust the temperature parameters of both components to fine-tune the model's learning process.

#########################################################################

Iteration 2: Reward Reflection:
We trained a RL policy using the provided reward function code and track the values of the individual components in the reward function as well as global policy metrics such as success rates and episode lengths after every 100 epochs and the maximum, mean, minimum values encountered:
velocity_reward: ['1.05', '6.33', '30.09', '66.72', '101.08', '133.44', '148.37', '175.90', '188.68', '204.21'], Max: 235.50, Mean: 116.81, Min: 1.05 
posture_reward: ['3.26', '3.31', '3.28', '3.27', '3.27', '3.25', '3.28', '3.28', '3.28', '3.27'], Max: 3.33, Mean: 3.27, Min: 3.02 
consecutive_successes: ['-0.05', '1.18', '2.31', '2.95', '3.20', '3.45', '3.52', '3.61', '3.66', '3.67'], Max: 3.79, Mean: 2.95, Min: -0.05 
episode_lengths: ['31.94', '85.50', '104.58', '123.72', '128.38', '130.28', '135.18', '138.20', '134.81', '136.41'], Max: 142.11, Mean: 120.44, Min: 25.89 
Please carefully analyze the policy feedback and provide a new, improved reward function that can better solve the task. Some helpful tips for analyzing the policy feedback:
    (1) If the success rates are always near zero, then you must rewrite the entire reward function
    (2) If the values for a certain reward component are near identical throughout, then this means RL is not able to optimize this component as it is written. You may consider
        (a) Changing its scale or the value of its temperature parameter
        (b) Re-writing the reward component 
        (c) Discarding the reward component
    (3) If some reward component magnitude is significantly larger, then you must re-scale its value to a proper range
Please analyze each existing reward component in the suggested manner above first, and then write the reward function code. The output of the reward function should consist of two items:
    (1) the total reward,
    (2) a dictionary of each individual reward component.
The code output should be formatted as a python code string: "```python ... ```".

Some helpful tips for writing the reward function code:
    (1) You may find it helpful to normalized the reward to a fixed range by applying transformations like torch.exp to the overall reward or its components
    (2) If you choose to transform a reward component, then you must also introduce a temeperature parameter inside the transformation function; this parameter must be a named variable in the reward function and it must not be an input variable. Each transformed reward component should have its own temperature variable
    (3) Make sure the type of each input variable is correctly specified; a float input variable should not be specified as torch.Tensor
    (4) most importantly, the reward code input variables must contain only attributes of the provided environment class definition (namely, variables that have prefix self.). Under no circumstance can you introduce new input variables.

#########################################################################

Iteration 3: Eureka Output:
Based on the provided feedback, it seems that the `posture_reward` component has little impact on the learning process, as its values are near identical throughout the training. I will modify the reward function to better balance the `velocity_reward` and `posture_reward` components, and also add a new reward component to encourage the agent to minimize the energy consumption of its action (i.e., `action_energy`).

Here is the updated reward function:

```python
import torch
from typing import Tuple, Dict

@torch.jit.script
def compute_reward(root_states: torch.Tensor, actions: torch.Tensor, dt: float) -> Tuple[torch.Tensor, Dict[str, torch.Tensor]]:
    # Calculate the forward velocity
    forward_velocity = root_states[:, 7]
    
    # Reward for forward velocity
    velocity_temperature = 5.0  # I increased the value from 1.0 to 5.0 to balance the impact of the velocity reward.
    velocity_reward = torch.exp(forward_velocity / velocity_temperature)
    
    # Calculate the height of the humanoid torso
    torso_height = root_states[:, 2]

    # Reward for maintaining upright posture
    posture_temperature = 10.0  # Increased the value from 1.0 to 10.0 to balance the impact of the posture reward.
    posture_reward = torch.exp(torso_height / posture_temperature)

    # Reward for minimizing energy consumption in actions
    action_energy = torch.norm(actions, p=2, dim=-1)
    energy_temperature = 5.0
    energy_reward = -torch.exp(action_energy / energy_temperature)  # Subtract this component to minimize energy consumption.
    
    # Combine rewards
    total_reward = velocity_reward + posture_reward + energy_reward
    
    reward_components = {'velocity_reward': velocity_reward, 'posture_reward': posture_reward, 'energy_reward': energy_reward}
    
    return total_reward, reward_components
```

I increased the temperature parameters for both the `velocity_reward` and `posture_reward` components to balance their impact on the learning process. Additionally, I added a new reward component, `energy_reward`, to encourage the agent to minimize the energy consumption in its actions. Experiment with different temperature parameter values if necessary to fine-tune the policy further.
\end{lstlisting}

\subsection{Negatively Correlated \ourmethod Reward Examples}
\label{appendix:negative-correlated-examples}
We display several negatively correlated but superior rewards that \ourmethod discovers in our experiments. 
\renewcommand\lstlistingname{Example}
\setcounter{lstlisting}{0}
\begin{lstlisting}[language=python, basicstyle=\fontfamily{\ttdefault}\scriptsize, caption={Task: ShadowHand, Human Normalized Score: 1.45, Correlation:$-0.26$}]
@torch.jit.script
def compute_reward(object_rot: torch.Tensor, goal_rot: torch.Tensor, fingertip_pos: torch.Tensor, object_pos: torch.Tensor) -> Tuple[torch.Tensor, Dict[str, torch.Tensor]]:
    # Compute the quaternion distance between the object's current orientation and the goal orientation
    q_dist = torch.sum((object_rot * goal_rot), dim=-1)
    q_dist = torch.min(q_dist, 1 - q_dist)  # Make sure the q_dist is in the range [0, 1]

    # Normalize the quaternion distance using a temperature parameter
    temp_rot = 0.5
    rot_reward = torch.exp(-temp_rot * q_dist)

    # Compute the distance between the fingertips and the object center
    fingertips_object_dist = torch.norm(fingertip_pos - object_pos[:, None], dim=-1)

    # Apply a threshold for the distance
    distance_threshold = 0.1
    close_enough = (fingertips_object_dist < distance_threshold).to(torch.float32)

    # Normalize the distance between fingertips and object center using an updated temperature parameter
    temp_dist = 10.0
    distance_reward = torch.mean(torch.exp(-temp_dist * fingertips_object_dist * close_enough), dim=-1)

    # Apply a penalty if the agent is not close enough to the object
    distance_penalty = 0.5 * (1 - torch.prod(close_enough, dim=-1))

    # Combine the reward components
    total_reward = rot_reward * distance_reward - distance_penalty

    # Store the reward components in a dictionary
    reward_components = {
        "rot_reward": rot_reward,
        "distance_reward": distance_reward,
        "distance_penalty": distance_penalty,
        "total_reward": total_reward,
    }

    return total_reward, reward_components
\end{lstlisting}
\begin{lstlisting}[language=python, basicstyle=\fontfamily{\ttdefault}\scriptsize, caption={Task: FrankaCabinet, Human Normalized Score: 11.98, Correlation:$-0.30$}]
@torch.jit.script
def compute_reward(franka_grasp_pos: torch.Tensor, drawer_grasp_pos: torch.Tensor, cabinet_dof_pos: torch.Tensor, 
                   franka_lfinger_pos: torch.Tensor, franka_rfinger_pos: torch.Tensor) -> Tuple[torch.Tensor, Dict[str, torch.Tensor]]:
    
    # Calculate the distance between the Franka grasping position and the cabinet grasping position
    grasp_distance = torch.norm(franka_grasp_pos - drawer_grasp_pos, dim=-1)

    # Calculate the distances between franke_lfinger_pos, franka_rfinger_pos and drawer_grasp_pos
    lfinger_distance = torch.norm(franka_lfinger_pos - drawer_grasp_pos, dim=-1)
    rfinger_distance = torch.norm(franka_rfinger_pos - drawer_grasp_pos, dim=-1)

    # Calculate the drawer opening distance
    drawer_opening = cabinet_dof_pos[:, 3]

    # Define temperature parameters for transforming the reward components
    grasp_distance_scaling = torch.tensor(20.0)
    handle_grasping_temperature = torch.tensor(20.0)
    drawer_opening_temperature = torch.tensor(20.0)

    # Transform the reward components
    grasp_distance_reward = 1.0 / (1.0 + grasp_distance_scaling * grasp_distance)
    handle_grasping_reward = torch.exp(-handle_grasping_temperature * (lfinger_distance + rfinger_distance))
    drawer_opening_reward = torch.exp(drawer_opening_temperature * drawer_opening)

    # Compute the total reward
    reward = grasp_distance_reward + handle_grasping_reward + drawer_opening_reward

    # Create a dictionary of individual reward components
    reward_components = {
        "grasp_distance_reward": grasp_distance_reward,
        "handle_grasping_reward": handle_grasping_reward,
        "drawer_opening_reward": drawer_opening_reward
    }

    return reward, reward_components
\end{lstlisting}

\subsection{\ourmethod from Human Initialization Examples}
\label{appendix:human-initialization-examples}
We display several examples of a single step in the \ourmethod from Human Initialization setting. In these examples, the first reward (Iteration 0) is the original human-written task reward, and Iteration 1 is the best reward after one step of \ourmethod improvement. 
\renewcommand\lstlistingname{Example}
\setcounter{lstlisting}{0}
\begin{lstlisting}[language=python, basicstyle=\fontfamily{\ttdefault}\scriptsize, caption={\ourmethod Human Initialization on Kettle. Human Success Rate: 0.11, \ourmethod Success Rate: 0.91}]
Iteration 1: Human Initialization:
```python
import torch

@torch.jit.script
def compute_reward(
    kettle_handle_pos, bucket_handle_pos, kettle_spout_pos,
    right_hand_ff_pos, right_hand_mf_pos, right_hand_rf_pos, right_hand_lf_pos, right_hand_th_pos,
    left_hand_ff_pos, left_hand_mf_pos, left_hand_rf_pos, left_hand_lf_pos, left_hand_th_pos,
):
    right_hand_finger_dist = (torch.norm(kettle_handle_pos - right_hand_ff_pos, p=2, dim=-1) + torch.norm(kettle_handle_pos - right_hand_mf_pos, p=2, dim=-1)
                            + torch.norm(kettle_handle_pos - right_hand_rf_pos, p=2, dim=-1) + torch.norm(kettle_handle_pos - right_hand_lf_pos, p=2, dim=-1) 
                            + torch.norm(kettle_handle_pos - right_hand_th_pos, p=2, dim=-1))
    left_hand_finger_dist = (torch.norm(bucket_handle_pos - left_hand_ff_pos, p=2, dim=-1) + torch.norm(bucket_handle_pos - left_hand_mf_pos, p=2, dim=-1)
                            + torch.norm(bucket_handle_pos - left_hand_rf_pos, p=2, dim=-1) + torch.norm(bucket_handle_pos - left_hand_lf_pos, p=2, dim=-1) 
                            + torch.norm(bucket_handle_pos - left_hand_th_pos, p=2, dim=-1))

    right_hand_dist_rew = right_hand_finger_dist
    left_hand_dist_rew = left_hand_finger_dist

    up_rew = torch.zeros_like(right_hand_dist_rew)
    up_rew = torch.where(right_hand_finger_dist < 0.7,
                    torch.where(left_hand_finger_dist < 0.7,
                                    0.5 - torch.norm(bucket_handle_pos - kettle_spout_pos, p=2, dim=-1) * 2, up_rew), up_rew)

    reward = 1 + up_rew - right_hand_dist_rew - left_hand_dist_rew
    reward_components = {
        'up_reward': up_rew,
        'right_hand_dist_reward': right_hand_dist_rew,
        'left_hand_dist_reward': left_hand_dist_rew,
    }

    return reward, reward_components
```

#########################################################################

Iteration 1: Reward Reflection:
We trained a RL policy using the provided reward function code and tracked the values of the individual components in the reward function as well as global policy metrics such as success rates and episode lengths after every 300 epochs and the maximum, mean, minimum values encountered:
up_reward: ['0.00', '0.00', '0.02', '-0.01', '0.04', '-0.02', '0.03', '-0.01', '0.02', '0.04'], Max: 0.12, Mean: 0.03, Min: -0.06 
right_hand_dist_reward: ['0.93', '0.34', '0.40', '0.38', '0.36', '0.44', '0.43', '0.42', '0.38', '0.42'], Max: 0.96, Mean: 0.41, Min: 0.27 
left_hand_dist_reward: ['1.47', '0.87', '0.69', '0.89', '0.83', '0.81', '0.83', '1.02', '0.84', '0.85'], Max: 3.57, Mean: 0.88, Min: 0.39 
consecutive_successes: ['0.00', '0.00', '0.00', '0.00', '0.00', '0.00', '0.00', '0.00', '0.00', '0.00'], Max: 0.01, Mean: 0.00, Min: 0.00 
episode_lengths: ['125.00', '122.45', '123.92', '124.24', '124.22', '124.83', '124.86', '124.42', '124.89', '124.18'], Max: 125.00, Mean: 123.81, Min: 111.56 
Please carefully analyze the policy feedback and provide a new, improved reward function that can better solve the task. Some helpful tips for analyzing the policy feedback:
    (1) If the success rates are always near zero, then you must rewrite the entire reward function
    (2) If the values for a certain reward component are near identical throughout, then this means RL is not able to optimize this component as it is written. You may consider
        (a) Changing its scale or the value of its temperature parameter
        (b) Re-writing the reward component 
        (c) Discarding the reward component
    (3) If some reward component magnitude is significantly larger, then you must re-scale its value to a proper range
Please analyze each existing reward component in the suggested manner above first, and then write the reward function code. The output of the reward function should consist of two items:
    (1) the total reward,
    (2) a dictionary of each individual reward component.
The code output should be formatted as a python code string: "```python ... ```".

Some helpful tips for writing the reward function code:
    (1) You may find it helpful to normalize the reward to a fixed range by applying transformations like torch.exp to the overall reward or its components
    (2) If you choose to transform a reward component, then you must also introduce a temperature parameter inside the transformation function; this parameter must be a named variable in the reward function and it must not be an input variable. Each transformed reward component should have its own temperature variable
    (3) Make sure the type of each input variable is correctly specified; a float input variable should not be specified as torch.Tensor
    (4) Most importantly, the reward code input variables must contain only attributes of the provided environment class definition (namely, variables that have prefix self.). Under no circumstance can you introduce new input variables.

#########################################################################

Iteration 2: Eureka Output:
From the feedback data you provided, it is evident that up_reward has a rather small value and success rate is always near zero. We can try some modifications to improve the reward function. First, we can update the distance rewards by adjusting their scale and applying an exponential transformation with an associated temperature. This will make their values range between 0 and 1. We will also slightly adjust the up_reward to get more value when the kettle and bucket are closer. Finally, we will add some weights to the reward components to balance their contribution to the total reward.

Here is the updated reward function:

```python
import torch

@torch.jit.script
def compute_reward(
    kettle_handle_pos, bucket_handle_pos, kettle_spout_pos,
    right_hand_ff_pos, right_hand_mf_pos, right_hand_rf_pos, right_hand_lf_pos, right_hand_th_pos,
    left_hand_ff_pos, left_hand_mf_pos, left_hand_rf_pos, left_hand_lf_pos, left_hand_th_pos,
):
    dist_temp = torch.tensor(0.1)
    
    right_hand_finger_dist = (torch.norm(kettle_handle_pos - right_hand_ff_pos, p=2, dim=-1) + torch.norm(kettle_handle_pos - right_hand_mf_pos, p=2, dim=-1)
                            + torch.norm(kettle_handle_pos - right_hand_rf_pos, p=2, dim=-1) + torch.norm(kettle_handle_pos - right_hand_lf_pos, p=2, dim=-1) 
                            + torch.norm(kettle_handle_pos - right_hand_th_pos, p=2, dim=-1))
    left_hand_finger_dist = (torch.norm(bucket_handle_pos - left_hand_ff_pos, p=2, dim=-1) + torch.norm(bucket_handle_pos - left_hand_mf_pos, p=2, dim=-1)
                            + torch.norm(bucket_handle_pos - left_hand_rf_pos, p=2, dim=-1) + torch.norm(bucket_handle_pos - left_hand_lf_pos, p=2, dim=-1) 
                            + torch.norm(bucket_handle_pos - left_hand_th_pos, p=2, dim=-1))

    right_hand_dist_rew = 1 - torch.exp(-dist_temp * right_hand_finger_dist)
    left_hand_dist_rew = 1 - torch.exp(-dist_temp * left_hand_finger_dist)

    up_rew = torch.zeros_like(right_hand_dist_rew)
    up_rew = torch.where(right_hand_finger_dist < 0.7,
                    torch.where(left_hand_finger_dist < 0.7,
                                    1 - torch.norm(bucket_handle_pos - kettle_spout_pos, p=2, dim=-1) * 2, up_rew), up_rew)

    reward = 5 * up_rew - 3 * right_hand_dist_rew - 2 * left_hand_dist_rew
    reward_components = {
        'up_reward': up_rew,
        'right_hand_dist_reward': right_hand_dist_rew,
        'left_hand_dist_reward': left_hand_dist_rew,
    }

    return reward, reward_components
```

Please test this updated reward function and provide new policy feedback data for further analysis if needed.
\end{lstlisting}
\lstinputlisting[language=python, basicstyle=\fontfamily{\ttdefault}\scriptsize, caption={\ourmethod Human Initialization on CatchUnderarm. Human Success Rate: 0.33, \ourmethod Success Rate: 0.64}]{appendix/eureka_examples/human_initialization/catch_underarm.py}

\subsection{\ourmethod from Human Reward Reflection}
\label{appendix:human-reward-reflection-example}
We display the raw dialogue that includes all human reward reflection texts as well as the generated \ourmethod rewards in our \ourmethod from human reward reflection experiment.
\renewcommand\lstlistingname{Example}
\setcounter{lstlisting}{0}
\lstinputlisting[language=python, basicstyle=\fontfamily{\ttdefault}\scriptsize, caption={\ourmethod from Human Reward Reflection}]{appendix/eureka_examples/human_reward_reflection/humanoid.py}

\subsection{\ourmethod and Human Reward Comparison}
\label{appendix:human-comparison-examples}
We display the human reward on a Dexterity task verbatim and contrast it with a \ourmethod-generated reward on the same task. As shown, the human reward is difficult to parse and has many commented-out blocks of reward components, suggesting history of trial-and-error reward design. In contrast, \ourmethod reward is clean and interpretable, amenable to post-hoc human inspection and editing. 

\renewcommand\lstlistingname{Example}
\setcounter{lstlisting}{0}
\lstinputlisting[language=python, basicstyle=\fontfamily{\ttdefault}\scriptsize, caption={Human reward for PushBlock}]{appendix/eureka_examples/human_comparison/push_block_human.py}
\begin{lstlisting}[language=python, basicstyle=\fontfamily{\ttdefault}\scriptsize, caption={Eureka reward for PushBlock}]
@torch.jit.script
def compute_reward(object_pos: Tensor,
                   left_hand_pos: Tensor,
                   right_hand_pos: Tensor,
                   left_goal_pos: Tensor,
                   right_goal_pos: Tensor) -> Tuple[Tensor, Dict[str, Tensor]]:

    # Temperature parameters for reward components
    temp_left_proximity: float = 0.5
    temp_right_proximity: float = 0.5
    temp_hand_distance: float = 0.1

    # Proximity reward for pushing the block to the left goal
    left_push_distance = torch.norm(object_pos - left_goal_pos, dim=1)
    left_proximity_reward = torch.exp(-temp_left_proximity * left_push_distance)

    # Proximity reward for pushing the block to the right goal
    right_push_distance = torch.norm(object_pos - right_goal_pos, dim=1)
    right_proximity_reward = torch.exp(-temp_right_proximity * right_push_distance)

    # Proximity rewards for the hands to be close to the block for better control
    left_hand_to_block = torch.norm(object_pos - left_hand_pos, dim=1)
    left_hand_reward = torch.exp(-temp_hand_distance * left_hand_to_block)

    right_hand_to_block = torch.norm(object_pos - right_hand_pos, dim=1)
    right_hand_reward = torch.exp(-temp_hand_distance * right_hand_to_block)

    # Final reward as a weighted sum of individual reward components
    reward = 0.25 * (left_proximity_reward + right_proximity_reward) + 0.25 * (left_hand_reward + right_hand_reward)

    reward_components = {
        "left_proximity_reward": left_proximity_reward,
        "right_proximity_reward": right_proximity_reward,
        "left_hand_reward": left_hand_reward,
        "right_hand_reward": right_hand_reward,
    }

    return reward, reward_components
\end{lstlisting}

\section{\arxiv{Limitations and Discussion}}
\arxiv{In this section, we discuss several limitations and discuss promising future work directions to address these limitations.}

\arxiv{One limitation is that \ourmethod currently is evaluated on simulation-based tasks with the exception of the preliminary real-robot experiment in App.~\ref{appendix:additional-results}, which demonstrates real-world hopping behavior learned via \ourmethod reward in a Sim2Real pipeline.
There are two promising directions in extending \ourmethod to the real-robot setting.
First, as demonstrated in App.~\ref{appendix:additional-results}, \ourmethod can be combined with Sim2Real approaches to first learn a policy in simulation and then transfer to the real-world~\citep{akkaya2019solving,kumar2021rma}; this is a well-established approach in the robotics community. Isaac Gym, in particular, has been used in many Sim2Real results~\citep{margolis2022rapid, handa2023dextreme}. Therefore, we believe Eureka's strong simulation results on various Isaac Gym tasks bodes well for real-world transfer.  Another approach is to instrument the real-world environments with sensors that can detect state measurements and then directly generate reward functions over these state variables, thereby enabling direct real-world reinforcement learning~\citep{gu2017deep, smith2022walk, yu2023language}. For either direction, progress in Sim2Real and state estimation will enhance the applicability of \ourmethod to the real-world setting. }

\arxiv{Another limitation is that \ourmethod requires a task fitness function $F$ to exist and easily definable by humans. While this assumption holds for a wide range of robot tasks (including all our benchmark tasks) and is required for reward design problem to be well-defined (Definition~\ref{def:rdp}), there are certain behavior that is hard to specify mathematically, such as ``running in a stable gait''. In those cases, we have presented \ourmethod from Human Feedback (Section~\ref{sec:eureka-human-feedback}) that uses human textual feedback to directly steer the reward generation, sidestepping having to specify $F$ that accurately captures the human intent. A promising future work direction is to use vision-language models (VLMs) to automatically provide textual feedback by feeding the policy videos into the VLMs~\citep{liu2023visual, zhu2023minigpt, yang2023dawn}.
}

\arxiv{Finally, our experiments currently take place on a single simulator and a fixed RL algorithm, with the exception of the Mujoco Humanoid experiment in App.~\ref{appendix:eureka-humanoid}. Given that \ourmethod does not take the identity of the simulation engine or RL algorithm as input, we believe that it should be straightforward to extend \ourmethod to custom simulators and optimizers. With regard to latter, it would be interesting to explore using model-predictive control methods~\citep{williams2015model, howell2022predictive}, which can potentially lead to faster behavior synthesis in terms of wall clock time.}

\end{document}